\documentclass[10pt]{article}

\usepackage[a4paper, top=2.0cm, bottom=2.0cm, left=2.1cm, right=2.1cm]{geometry}
\usepackage{amsmath, amssymb}
\usepackage{graphicx}

\usepackage[T1]{fontenc}
\usepackage[hidelinks,colorlinks=false]{hyperref}
\usepackage{booktabs}
\usepackage{tabularx}
\usepackage{caption}
\usepackage{xcolor}
\usepackage{multicol}
\usepackage{authblk}
\usepackage{lineno}
\usepackage{setspace}
\usepackage{fancyhdr}
\usepackage{helvet}

\graphicspath{{./}}

\newcommand{\AUCpt}{\textsc{auc}\,pp}

\newcommand{\nCells}[1]{%
  \ifx&#1&%
    \PackageError{cell_counts}{empty key}%
      {Specify a key like \nCells{llmtool_evaluable}}%
  \else%
    \ifcsname nCells@#1\endcsname%
      \csname nCells@#1\endcsname%
    \else%
      \PackageError{cell_counts}{unknown key '#1'}%
        {Add row '#1' to frozen_v3/cell_counts_taxonomy.csv and re-run scripts/build/build_cell_counts_sty.py.}%
    \fi%
  \fi%
}

\expandafter\newcommand\csname nCells@cartesian_total\endcsname{93}  
\expandafter\newcommand\csname nCells@llmtool_evaluable\endcsname{81}  
\expandafter\newcommand\csname nCells@llmtool_combined_reference\endcsname{90}  
\expandafter\newcommand\csname nCells@llmtool_per_tool_primary\endcsname{27}  
\expandafter\newcommand\csname nCells@mimic_cells\endcsname{36}  
\expandafter\newcommand\csname nCells@llmtool_mimic\endcsname{36}  
\expandafter\newcommand\csname nCells@llmtool_mimic_primary\endcsname{36}  
\expandafter\newcommand\csname nCells@ehrshot_cells\endcsname{39}  
\expandafter\newcommand\csname nCells@llmtool_ehrshot_primary\endcsname{39}  
\expandafter\newcommand\csname nCells@nhanes_cells\endcsname{6}  
\expandafter\newcommand\csname nCells@llmtool_nhanes\endcsname{6}  
\expandafter\newcommand\csname nCells@llmtool_nhanes_primary\endcsname{6}  
\expandafter\newcommand\csname nCells@llmtool_nhanes_sensitivity\endcsname{9}  
\expandafter\newcommand\csname nCells@nhanes_primary_diseases\endcsname{2}  
\expandafter\newcommand\csname nCells@nhanes_coarse_proxy_diseases\endcsname{3}  
\expandafter\newcommand\csname nCells@pak_nhanes\endcsname{6}  
\expandafter\newcommand\csname nCells@audit_class_audit_ok\endcsname{1}  
\expandafter\newcommand\csname nCells@audit_class_terminate_early\endcsname{1}  
\expandafter\newcommand\csname nCells@audit_class_terminate_late\endcsname{1}  
\expandafter\newcommand\csname nCells@audit_class_topup_needed\endcsname{10}  
\expandafter\newcommand\csname nCells@precision_k_rows\endcsname{31}  
\expandafter\newcommand\csname nCells@disease_panel\endcsname{13}  
\expandafter\newcommand\csname nCells@audit_total\endcsname{13}  
\expandafter\newcommand\csname nCells@audit_topup_needed\endcsname{10}  
\expandafter\newcommand\csname nCells@audit_ok\endcsname{1}  
\expandafter\newcommand\csname nCells@audit_finalize_strict\endcsname{3}  
\expandafter\newcommand\csname nCells@nhanes_labeled_icds\endcsname{5}  
\expandafter\newcommand\csname nCells@nhanes_unlabeled_icds\endcsname{8}  
\expandafter\newcommand\csname nCells@external_cohorts\endcsname{3}  
\expandafter\newcommand\csname nCells@mcp_tools\endcsname{5}  
\expandafter\newcommand\csname nCells@external_llm_tools\endcsname{3}  

\title{Agentic discovery of blood biomarkers \\ from distilled private health records}
\author[1,2,3,*]{Seffi Cohen}
\author[1,2,3]{Liat Antwarg Friedman}
\author[3]{Amir Anisman}
\author[1,2]{Ruth Johnson}
\author[1,2]{Michelle M.~Li}
\author[1]{Ayush Noori}
\author[1,2]{Ben Reis}
\author[2,3,5]{Ran Balicer}
\author[2,3,4,$\dagger$]{Noa Dagan}
\author[1,$\dagger$]{Marinka Zitnik}
\affil[1]{Department of Biomedical Informatics, Harvard Medical School, Boston, MA, USA}
\affil[2]{The Ivan and Francesca Berkowitz Family Living Laboratory Collaboration at Harvard Medical School and Clalit Research Institute, Boston, MA, USA}
\affil[3]{Clalit Research Institute, Innovation Division, Clalit Health Services, Ramat-Gan, Israel}
\affil[4]{Computer and Information Science, Ben Gurion University of the Negev, Be'er Sheva, Israel}
\affil[5]{Department of Epidemiology, Biostatistics and Community Health Sciences. Ben-Gurion University of the Negev, Be'er Sheva, Israel}
\affil[$\dagger$]{These authors contributed equally to this work.}
\affil[*]{Corresponding author: Seffi Cohen, \texttt{seffi\_cohen@hms.harvard.edu}}
\date{}

\begin{document}

\maketitle

\begin{abstract}
Routine complete blood counts (CBCs) could yield new biomarkers, but the private records needed to evaluate candidates cannot be shared with frontier language model agents that excel at discovery. We distilled the evidence held in the Clalit Health Services panel of over 5.4 million patients into a released scoring tool: for each of 13 immune-mediated diseases, a graph attention network was trained inside the data boundary to predict the case-control AUC of candidate CBC expressions, and only the trained weights were released. The tool grounds an agent's propose-score-refine loop in real-world data without exposing any patient data. In external validation, agent-discovered expressions improved on their literature-seeded starting points by a median of 4.18 AUC percentage points, and across three independent cohorts, reranking the candidates of three frontier research tools improved on their first choices in most comparisons, with gains that varied by cohort. The released scorer supports privacy-preserving biomarker hypothesis generation.
\end{abstract}

\section{Introduction}\label{sec:intro}
The complete blood count is one of the most widely used laboratory tests in routine care, and simple arithmetic summaries of its components, including the neutrophil-to-lymphocyte ratio (NLR), red cell
distribution width and mean platelet volume remain attractive biomarkers because they are inexpensive, interpretable, and broadly available across health systems \cite{zahorec2001nlr,templffigeton2014nlr,buonacera2022nlr,felker2007rdw,mpvinflammation}.
Manually enumerating all possible blood biomarker expressions is intractable because there are ~26 trillion combinations 
Fortunately, new blood test biomarkers can also be discovered through AI methods such as agentic research frameworks, which utilize skills and tools to review the literature, propose new hypotheses, and design and execute research experiments  \cite{boiko2023autonomous,mbran2024chemcrow,gottweis2026coscientist}. A large language model (LLM) agent can propose candidate expressions based on its parametric memory or by retrieving and reasoning over related literature. However, the development of candidate expressions could be improved by grounding proposals in real-world large-scale patient-level medical data rather than broad scientific literature, enabling evaluation and continuous refinement of future proposed expressions.
The cohorts most informative for evaluating candidate biomarkers are private electronic health records (EHRs) held behind institutional firewalls \cite{rajkomar2018ehr,topol2019dl} or air-gapped networks. Importantly, frontier AI models and agents, such as Gemini~3~Pro \cite{gemini3pro2024}, GPT Deep Research
\cite{gpt_deep_research2024}, and SciSpace BM Agent \cite{scispace2024},
can propose expressions from public literature, but often cannot be deployed inside protected research environments to train on real-world patient data.
Federated learning can train shared models across protected institutional datasets without centralizing patient-level data \cite{mcmahan2017fedavg,kaissis2020fl}, but it is designed primarily for collaborative model training rather than iterative evaluation of candidate biomarker proposals by frontier models with real-time feedback. What is missing is an artifact that learns from private clinical data inside a private network and can then rank new CBC
expressions outside the private network without querying patient records.

\begin{figure}[!t]
\centering
\includegraphics[width=\textwidth]{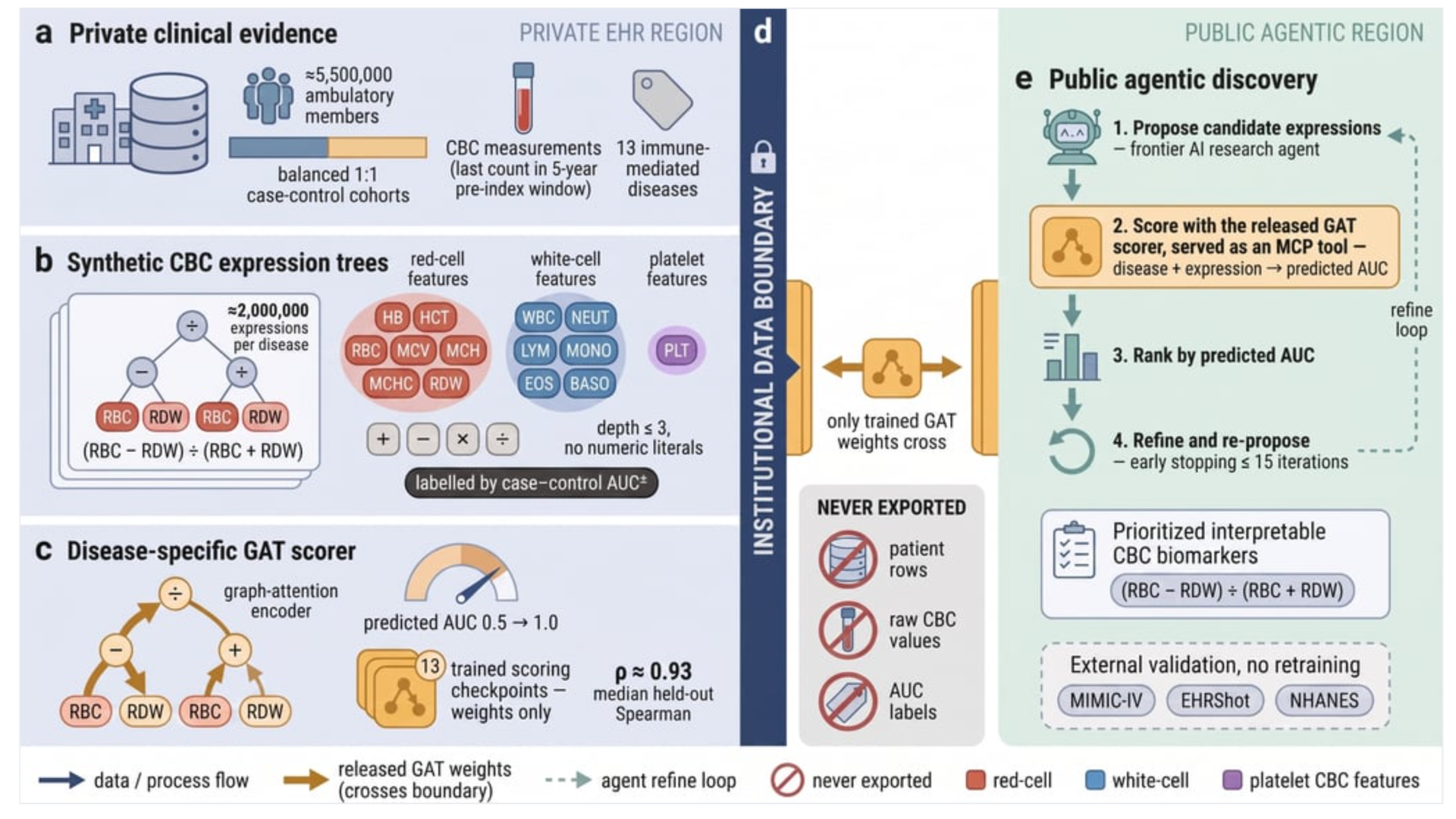}
\caption[Privacy-preserving public scoring framework for CBC biomarker discovery.]{
\textbf{Privacy-preserving public scoring framework for CBC biomarker discovery.}
\textbf{A)} Private clinical evidence remains inside the Clalit data boundary:
anonymized EHR-derived cohorts, CBC measurements, and disease labels are used
only within the protected environment.
\textbf{B)} Synthetic CBC expression trees are generated from CBC variables and
arithmetic operators, with leaves restricted to measured CBC features and no
numeric literals.
\textbf{C)} Disease-specific graph-attention networks learn to predict
expression-level AUC from expression structure, producing compact scoring
checkpoints.
\textbf{D)} Only trained GAT weights are exported; patient rows, raw CBC values,
and expression-level AUC labels remain locked inside the institutional boundary.
\textbf{E)} A public frontier artificial intelligence (AI) agent uses the released scorer in a
propose--score--rank--refine loop to prioritize interpretable CBC biomarker
expressions.
}
\label{fig:overview}
\end{figure}

We address this gap by distilling a private cohort into a released,
patient-level-free scorer for complete blood count (CBC) expression trees. Inside the Clalit
Health Services ambulatory member panel of 5{,}437{,}870 patients, we synthesize 2 million candidates CBC arithmetic expressions per disease. Each candidate biomarker is a binary tree of depth at most three over a 14-feature CBC vocabulary and the operator set $\{+,-,\times,\div\}$, with numeric literals forbidden. Within the protected environment, every expression is scored against a disease label by the direction-agnostic univariate area under the receiver operating characteristic curve (AUC). We then train a disease-specific graph attention model (GAT) \cite{velickovic2018gat,brody2022gatv2} to predict that AUC and expose this GAT model as a tool via the model context protocol (MCP) \cite{anthropic2024mcp}. This makes it possible for a frontier commercial language model search agent \cite{anthropic2026claude}
to propose, score, and refine candidate biomarkers while interacting only with the released GAT scorer (Figure~\ref{fig:overview}). 

We evaluate the scorer in two complementary settings. First, we use it
as the ranking signal in an iterative LLM-guided expression search loop
across 13 immune-related diseases. This search benchmark measures whether the released
scorer can guide candidate generation: the loop observes only predicted AUC
scores, and early stopping is determined only by the predicted score
trajectory. After the loop terminates, the starting expression and the
selected expression are evaluated by the validation pipeline to obtain
post-hoc realized AUCs. These realized AUCs are not exposed to the agent
and do not affect ranking or stopping. Relative to a fixed starting
expression, the selected expression improves post-hoc realized AUC by a
median of $+4.18$\, AUC percentage points (\AUCpt{}) across diseases. Second, we test whether the scorer improves the ordering
of expressions produced by external LLM agentic tools. Candidate expressions are re-scored without
retraining in three independent cohorts:
The Medical Information  Mart for Intensive Care (MIMIC-IV)
\cite{johnson2023mimic}, EHRShot \cite{wornow2023ehrshot}, and the National Health and Nutrition Examination 
Survey (NHANES) \cite{cdc1988nhanes}. In this tool re-ranking benchmark, the scorer's
top-ranked expression outperforms the tool's first expression in 67.9\%  [57.1\%,77.1\%] 
of 81 comparisons, with a median
gain of $+1.50$\,\AUCpt{}; adding the 9 NHANES comparisons, which we report only as a sensitivity
panel, gives 71.1\% of 90.
External transfer is cohort-dependent \cite{finlayson2021shift,subbaswamy2020shift,chekroud2024illusory}, and we therefore treat these
results as evidence for prioritization rather than standalone screening.

The main contributions are:  
\begin{itemize}
    \item (i) A privacy-preserving distillation framework that converts evidence from patient-level EHR data into a portable scoring model that can evaluate new biomarker hypotheses outside the institutional data boundary.
    \item (ii) A disease-specific GAT scorer for CBC biomarker discovery across 13 immune-mediated diseases, released as an open-source resource.
    \item (iii) A generalizable paradigm for connecting frontier AI agents to otherwise inaccessible clinical evidence.
\end{itemize}

\section{Results}\label{sec:results}

\subsection{A private cohort was distilled into a released scorer for CBC expression trees}
\begin{figure}[!ht]
\centering
\includegraphics[width=\textwidth]{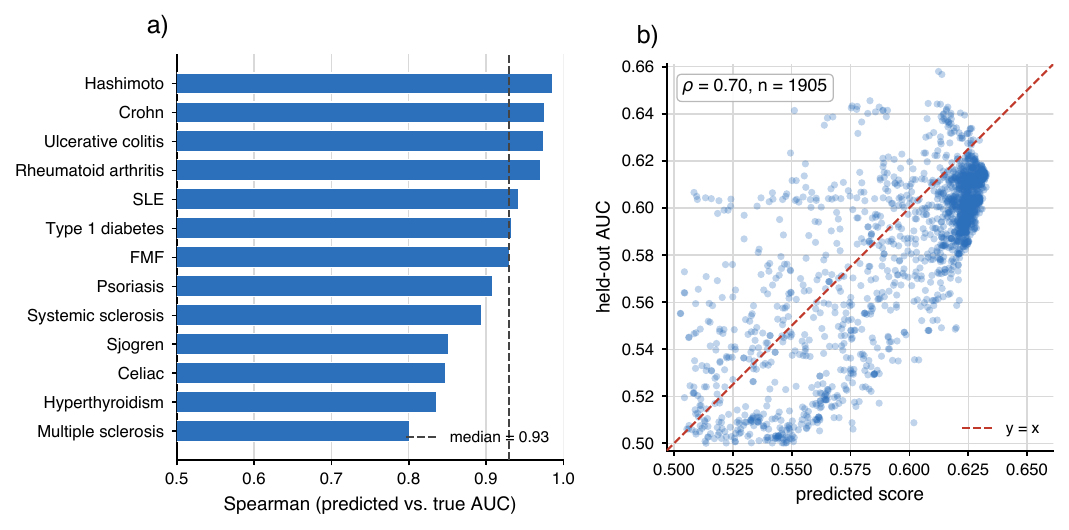}
\caption{Scorer-training agreement. a)~Per-disease Spearman correlation between the scorer's predicted AUC and the ground-truth AUC on the held-out synthetic-expression split. b)~Predicted score versus the ground-truth AUC for the agent-proposed candidate expressions of a representative disease (rheumatoid arthritis).}
\label{fig:training_agreement}
\end{figure}

The exported model in this study is a trained scoring Graph Attention Network (GAT), trained on expression--AUC pairs rather than on patient-level records. For each of 13 immune-mediated diseases, we trained a disease-specific GAT inside the Clalit Health Services (CHS) data boundary. The source panel comprised approximately 5{,}500{,}000
ambulatory members, from which balanced 1:1 case/control subcohorts were drawn for each disease as detailed in Table~\ref{tab:cohort_sizes}. For each disease, the training corpus contained approximately 2{,}000{,}000 synthetic CBC arithmetic-expression trees. Each tree used a 14-feature CBC
vocabulary: hemoglobin (HB), hematocrit (HCT), red blood cell count (RBC), white blood cell count (WBC), platelet count (PLT), mean corpuscular volume (MCV), mean corpuscular hemoglobin (MCH), mean corpuscular hemoglobin concentration (MCHC), red cell distribution width (RDW), and the neutrophil, lymphocyte, monocyte, eosinophil, and basophil percentages (NEUTpct or NEUT\%, LYMpct or LYM\%, and so on) and the  operator set $\{+,-,\times,\div\}$, with a maximum depth of three and no numeric literals. Inside CHS, each expression was labeled by its direction-agnostic AUC for separating cases from controls. The scorer was then trained to predict that AUC from expression structure alone as illustrated in Figure~\ref{fig:overview}.
On a held-out split of the per-disease synthetic-expression corpus, the scorer predicts AUC ordering closely: Across the 13 diseases, the rank correlation is high and consistent - the median Spearman rank correlation between predicted and ground truth AUC was $\approx 0.93$ (range $0.80$, multiple sclerosis, to $0.99$, Hashimoto thyroiditis), the mean absolute error (MAE) of the predicted AUC on the $[0.5,1.0]$ scale did not exceed $0.019$ (maximum $0.0186$, systemic lupus erythematosus), and the normalised discounted cumulative gain over the top 5{,}000 ranked held-out expressions (NDCG@5000) was 0.984--0.999 as detailed in Table~\ref{tab:training_diag} and Figure~\ref{fig:training_agreement}.

\begin{table}[!ht]
\centering
\small
\caption{Cohorts and disease panel.}
\label{tab:cohort_sizes}
\begin{tabular}{llrrrr}
\toprule
ICD & Disease & $n_{\mathrm{Clalit}}$ & $n_{\mathrm{MIMIC}}$ & $n_{\mathrm{EHRShot}}$ & $n_{\mathrm{NHANES}}$ \\
\midrule
242 & Hyperthyroidism & 6,138 (3,069) & 73,171 (233) & 3,308 (51) & \textit{49,510 (4,808)}\textsuperscript{\textdagger} \\
250 & Type 1 Diabetes (T1D) & 25,314 (12,657) & 72,140 (306) & 3,186 (88) & \textit{52,231 (6,057)}\textsuperscript{\textdagger} \\
277 & Familial Mediterranean Fever (FMF) & 9,533 (4,766) & 29,929 (0) & 2,568 (3) & \textit{n.a.} \\
340 & Multiple Sclerosis (MS) & 5,960 (2,980) & 72,712 (126) & 2,815 (14) & \textit{n.a.} \\

555 & Crohn's Disease & 20,513 (10,256) & 71,091 (243) & 2,591 (30) & \textit{n.a.} \\
556 & Ulcerative Colitis (US) & 20,385 (10,192) & 69,635 (210) & 2,490 (47) & \textit{n.a.} \\
696 & Psoriasis & 111,835 (55,917) & 73,585 (313) & 2,859 (69) & 16,997 (464) \\
714 & Rheumatoid Arthritis (RA) & 61,188 (30,594) & 71,878 (463) & 2,769 (93) & 38,887 (2,655) \\

2452 & Hashimoto Thyroiditis & 54,382 (27,191) & 75,883 (94) & 3,327 (24) & \textit{49,510 (4,808)}\textsuperscript{\textdagger} \\
5790 & Celiac Disease & 8,341 (4,170) & 70,115 (100) & 2,935 (59) & \textit{n.a.} \\
7100 & Systemic Lupus Erythematosus (SLE) & 6,024 (3,012) & 69,662 (138) & 2,635 (118) & \textit{n.a.} \\
7101 & Systemic Sclerosis (SSc) & 2,390 (1,195) & 68,113 (38) & 2,708 (20) & \textit{n.a.} \\

7102 & Sjögren's Syndrome & 4,128 (2,064) & 72,990 (82) & 2,901 (36) & \textit{n.a.} \\
\bottomrule
\end{tabular}
\begin{flushleft}\footnotesize Per-disease Clalit $n$ are total subjects after 1:1 case/control balancing. External-cohort $n$ is the total number of subjects after preprocessing. \textit{n.a.} = no harmonized NHANES questionnaire item maps to this ICD, so the disease cannot be evaluated on NHANES at all (eight diseases: 277, 340, 555, 556, 5790, 7100, 7101, 7102). \textsuperscript{\textdagger}\,Italicised NHANES $n$ marks the three phenotypes (242, 250, 2452) for which an NHANES label does exist but rests on a coarse self-report questionnaire item rather than a verified diagnosis; these are excluded from the primary NHANES analysis and are reported only as a sensitivity analysis. The primary NHANES analysis therefore uses the two verified-diagnosis phenotypes, Psoriasis (696, $n = 16{,}997$) and Rheumatoid arthritis (714, $n = 38{,}887$).\end{flushleft}
\end{table}

\begin{table}[!ht]
\centering
\small
\caption{Per-disease scorer-training diagnostics on the held-out synthetic-expression split.}
\label{tab:training_diag}
\begin{tabular}{llrrrr}
\toprule
\textbf{ICD} & \textbf{Disease} & \textbf{Epochs} & \textbf{Val Spearman} & \textbf{Val MAE(AUC)} & \textbf{NDCG@5000} \\
\midrule
242 & Hyperthyroidism & 67 & 0.836 & 0.0133 & 0.998 \\
250 & Type 1 diabetes & 106 & 0.933 & 0.0055 & 0.998 \\
277 & Familial Mediterranean fever & 88 & 0.930 & 0.0078 & 0.999 \\
340 & Multiple sclerosis & 90 & 0.801 & 0.0096 & 0.991 \\
555 & Crohn's disease & 109 & 0.975 & 0.0072 & 0.997 \\
556 & Ulcerative colitis & 173 & 0.974 & 0.0071 & 0.999 \\
696 & Psoriasis & 26 & 0.908 & 0.0046 & 0.993 \\
714 & Rheumatoid arthritis & 176 & 0.971 & 0.0041 & 0.999 \\
2452 & Hashimoto thyroiditis & 129 & 0.986 & 0.0030 & 0.999 \\
5790 & Celiac disease & 81 & 0.848 & 0.0142 & 0.999 \\
7100 & Systemic lupus erythematosus & 67 & 0.942 & 0.0186 & 0.997 \\
7101 & Systemic sclerosis & 112 & 0.894 & 0.0176 & 0.997 \\
7102 & Sjögren's syndrome & 91 & 0.851 & 0.0164 & 0.984 \\
\bottomrule
\end{tabular}
\begin{flushleft}\footnotesize Val Spearman, Val MAE of the AUC, and NDCG@5000 are computed on the held-out synthetic-expression split per disease scorer; they characterize how faithfully the trained scorer reproduces the true held-out AUC ordering of candidate expressions. \end{flushleft}
\end{table}

The scoring model, a GAT model trained, per disease, to predict the AUC of CBC-based expressions from its parsed structure, does not hold any patient-level data and can thus be released as an open-source resource. Patient rows, individual CBC measurements, and expression-level CHS AUC labels remain inside the institutional environment. After expressions are proposed by a discovery agent, the scorer can rank candidate biomarker expressions in a way that reflects the EHR data. In other words, at inference, the released scorer accepts a disease identifier and an expression and returns the predicted-AUC score in $[0.5,1.0]$. This design makes the scorer usable by any external agent as a ranking resource while avoiding patient-row access at inference.

\subsection{Agent-discovered expressions did not uniformly generalize across three external validation cohorts}

We first used the released scorer as the ranking signal in an iterative search loop. A language model research agent (Opus 4.7 DeepResearch) proposed CBC expression candidates, received predicted AUC scores from our GAT tool, and used the highest predicted AUC expressions to seed the next round. Each disease run
therefore produced one discovered expression: the highest predicted AUC expression identified during the loop. Measured validation AUC was computed only after candidate selection and was not used for ranking or early stopping.

To test whether the discovered expressions generalized beyond CHS, we rescored each discovered expression and its starting expression on three external cohorts, without retraining: MIMIC-IV \cite{johnson2023mimic}
(mean $n = 71{,}748$ critical-care patients per evaluable disease; range $68{,}113$--$75{,}883$), EHRShot \cite{wornow2023ehrshot} (mean $n = 2{,}853$ Stanford ambulatory patients per disease; range $2{,}490$--$3{,}327$), and NHANES \cite{cdc1988nhanes} (cross-sectional participants; $n = 16{,}997$ for psoriasis and $n = 38{,}887$ for rheumatoid arthritis, the two verified-diagnosis phenotypes) with 500-resample percentile bootstrap 95\% CIs confidence intervals (CIs).

\begin{figure}[!ht]
\centering
\includegraphics[width=\textwidth]{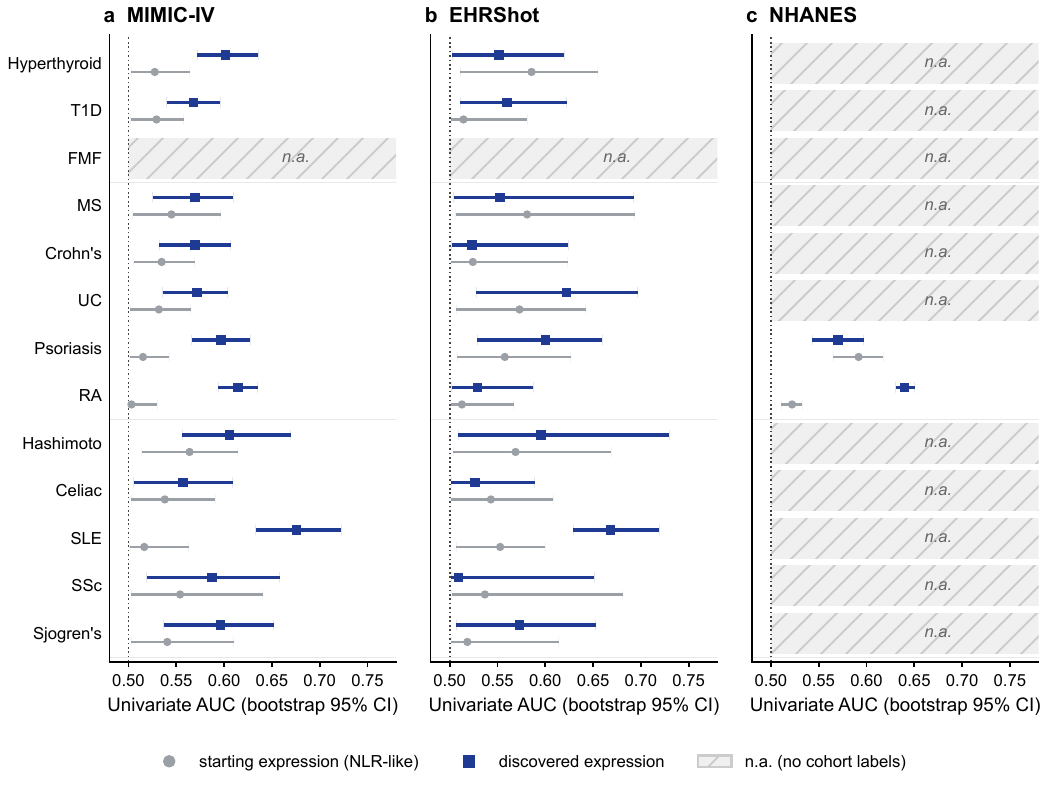}
\caption[External validation across MIMIC-IV, EHRShot, and NHANES.]{
\textbf{External validation of the starting expression and the discovered
expression on three independent cohorts.} \textbf{a)}~MIMIC-IV.
\textbf{b)}~EHRShot. \textbf{c)}~NHANES, restricted to the two
verified-diagnosis phenotypes. Grey circles mark the starting expression and
blue squares the discovered expression; horizontal bars are bootstrap 95\%
confidence intervals, and hatched rows (\textit{n.a.}) mark cells without
usable cohort labels.}
\label{fig:external_forest}
\end{figure}

External validation was strongest on MIMIC-IV. The discovered expression's point AUC matched or exceeded the starting expression's point AUC in 12 of 12 evaluable MIMIC-IV disease cells; familial Mediterranean fever (FMF) was excluded from
AUC evaluation because MIMIC-IV holds no FMF cases. Six of the 12 MIMIC-IV gains also satisfied the stricter criterion that the discovered expression's 95\% CI lower bound exceeded the starting expression's point AUC as shown in Table~\ref{tab:external_winner} and Figure~\ref{fig:external_forest}a.
The largest MIMIC-IV gain was for Systemic lupus erythematosus (SLE), where the discovered expression reached AUC $0.676$ compared with $0.516$ for the Opus 4.7 research agent without the GAT scoring tool ($\mathrm{NEUTpct}/\mathrm{LYMpct}$).

\begin{table}[!ht]
\centering
\scriptsize
\caption{External validation per disease and cohort.}
\label{tab:external_winner}
\resizebox{\textwidth}{!}{%
\begin{tabular}{llll}
\toprule
Disease & MIMIC-IV & EHRShot & NHANES \\
\midrule
Hyperthyroidism & 0.527 [0.503, 0.564] / 0.602 [0.572, 0.636] & 0.586 [0.511, 0.655] / 0.551 [0.503, 0.620] & \textit{n.a.} \\
Type 1 Diabetes & 0.529 [0.503, 0.558] / 0.568 [0.540, 0.596] & 0.514 [0.502, 0.581] / 0.560 [0.511, 0.622] & \textit{n.a.} \\
Multiple Sclerosis & 0.545 [0.504, 0.596] / 0.570 [0.525, 0.610] & 0.581 [0.506, 0.694] / 0.553 [0.504, 0.693] & \textit{n.a.} \\
Crohn's Disease & 0.535 [0.505, 0.569] / 0.570 [0.532, 0.607] & 0.524 [0.501, 0.623] / 0.523 [0.502, 0.624] & \textit{n.a.} \\
Ulcerative Colitis & 0.532 [0.502, 0.565] / 0.572 [0.536, 0.604] & 0.573 [0.507, 0.642] / 0.622 [0.528, 0.697] & \textit{n.a.} \\
Psoriasis & 0.515 [0.501, 0.543] / 0.597 [0.566, 0.627] & 0.558 [0.508, 0.627] / 0.600 [0.529, 0.660] & 0.592 [0.565, 0.617] / 0.570 [0.543, 0.597] \\
Rheumatoid Arthritis & 0.503 [0.500, 0.530] / 0.615 [0.594, 0.635] & 0.513 [0.501, 0.567] / 0.529 [0.502, 0.587] & 0.522 [0.511, 0.533] / 0.640 [0.630, 0.651] \\
Hashimoto Thyroiditis & 0.564 [0.514, 0.614] / 0.606 [0.556, 0.670] & 0.569 [0.504, 0.669] / 0.596 [0.509, 0.730] & \textit{n.a.} \\
Celiac Disease & 0.538 [0.503, 0.591] / 0.557 [0.505, 0.609] & 0.543 [0.501, 0.608] / 0.526 [0.502, 0.589] & \textit{n.a.} \\
Systemic Lupus Erythematosus & 0.516 [0.501, 0.563] / 0.676 [0.633, 0.723] & 0.553 [0.507, 0.600] / 0.668 [0.629, 0.719] & \textit{n.a.} \\
Systemic Sclerosis & 0.554 [0.503, 0.641] / 0.588 [0.519, 0.658] & 0.537 [0.502, 0.681] / 0.509 [0.501, 0.651] & \textit{n.a.} \\
Sjögren's Syndrome & 0.541 [0.503, 0.610] / 0.596 [0.537, 0.652] & 0.518 [0.502, 0.614] / 0.573 [0.507, 0.653] & \textit{n.a.} \\
\bottomrule
\end{tabular}%
}
\end{table}

On EHRShot, transfer was weaker: the discovered expression exceeded the starting expression's point AUC in 7 of 12 evaluable comparisons (FMF is omitted here because its three EHRShot positives preclude stable AUC estimation), but only 1 of 12 cleared the strict CI-lower-bound criterion, as shown in Figure~\ref{fig:external_forest}b. The NHANES primary analysis was restricted to the two verified-diagnosis phenotypes with sufficient case counts for bootstrap inference: psoriasis ($n=16{,}997$) and Rheumatoid arthritis (RA) ($n=38{,}887$). In NHANES, the RA expression improved from AUC $0.522$ to $0.640$ with separated CIs, whereas the psoriasis expression regressed from AUC $0.592$ to $0.570$ with overlapping CIs, as shown in Figure~\ref{fig:external_forest}c. Thus, the expressions discovered through the private-cohort scorer transferred well in some disease--cohort settings, but did not uniformly generalize across external case definitions and population settings.

We also evaluated whether the discovered expressions could enrich cases in the top-ranked tail of an external cohort. Patients were ranked by absolute deviation from the cohort median of the expression value, and precision-at-K was computed at $K \in \{0.5\%,1\%,5\%\}$ within each cohort separately. Across the 27 evaluable cohort--disease cells at $K=1\%$, enrichment over base prevalence reached $6.38\times$, but only 15 of the 27 cells exceeded random screening at that threshold and 9 achieved no enrichment at all. The largest enrichments were Hashimoto thyroiditis on MIMIC-IV ($6.38\times$), Hashimoto thyroiditis on EHRShot ($4.20\times$), and SLE on MIMIC-IV ($3.62\times$), as shown in Table~\ref{tab:precision_at_k}. Enrichment is therefore highly uneven across cohorts for the same expression, and this metric should be read as a screening-tail diagnostic rather than as evidence of deployable screening performance.

\subsection{The released GAT improved the ranking of leading LLM-generated biomarker candidates}

We next tested whether the released scorer could improve the ranking of candidate expressions proposed by independent LLM biomarker tools. Three frontier agentic tools: GPT Deep Research, Gemini 3 Pro, and SciSpace BM Agent, each produced approximately 200 candidate CBC arithmetic expressions per disease (range 168--229), using a prompt optimized for this task with the Genetic Algorithm Applied to Prompt Optimization (GAAPO) \cite{secheresse2025gaapo}. This evaluation did not use our iterative search agent. Instead, we compared two expressions from the same produced candidate set: the tool's own first produced expression and the expression ranked highest by the released GAT scorer. Both were then evaluated on the corresponding external cohort.

The primary benchmark comprises \nCells{llmtool_evaluable} verified diagnosis comparisons (Methods); the three MIMIC-IV $\times$ FMF comparisons were unevaluable because that cohort holds no FMF cases at all. Across the primary panel, the scorer-ranked expression outperformed the tool's own first expression in 67.9\% [57.1\%,77.1\%] of comparisons (55 of \nCells{llmtool_evaluable}), with a median paired gain of $+1.50$\,\AUCpt{} (mean $+1.81$\,\AUCpt{}; paired Wilcoxon signed-rank $p = 4.0\times10^{-5}$).

The advantage was strongly cohort-dependent. On MIMIC-IV, the scorer-ranked expression won 86.1\% of \nCells{llmtool_mimic} comparisons, with a median paired gain of $+3.67$\,\AUCpt{} ($p = 1.9\times10^{-5}$). On the NHANES verified-diagnosis subcohort, restricted to psoriasis and RA, the scorer-ranked expression won 83.3\% of \nCells{llmtool_nhanes_primary} comparisons with median $\Delta = +3.83$\,\AUCpt{}; with only \nCells{llmtool_nhanes_primary} comparisons, however, the paired test does not reach conventional significance ($p = 0.063$), so this sub-panel suggests, but cannot establish, a reranking advantage on NHANES. By contrast, EHRShot showed no detectable reranking advantage: across \nCells{ehrshot_cells} comparisons the scorer-ranked expression won 48.7\%, with a median $\Delta$ of $0.00$\,\AUCpt{} and no evidence against the paired null ($p = 0.92$).

The direction of the effect reproduced across all three tools within the primary panel, each contributing \nCells{llmtool_per_tool_primary} comparisons: GPT Deep Research 74.1\% scorer wins (median $\Delta = +1.25$\,\AUCpt{}), Gemini 3 Pro 66.7\% (median $+1.54$\,\AUCpt{}), and SciSpace BM Agent 63.0\% (median $+1.05$\,\AUCpt{}).

The \nCells{llmtool_nhanes_sensitivity} NHANES comparisons whose labels rest on a coarse self-report questionnaire item rather than a verified diagnosis (hyperthyroidism, type~1 diabetes, and Hashimoto thyroiditis) were held out of the primary panel and are reported here only as a labelled sensitivity analysis. In that sub-panel the scorer-ranked expression won all \nCells{llmtool_nhanes_sensitivity} comparisons, with a median paired gain of $+8.94$\,\AUCpt{} (mean $+7.25$\,\AUCpt{}). Pooling the sensitivity cells with the primary panel would raise the overall figure to 71.1\% of \nCells{llmtool_combined_reference} comparisons with a median gain of $+1.91$\,\AUCpt{}; because those labels are the weakest in the panel and inflate the estimate, we report that combined figure for reference only and do not treat it as the headline result.

Overall, the external evaluation indicates that the private-cohort scorer can enhance the ranking of external LLM-tool candidates, with the caveat that the gain is carried by MIMIC-IV, is suggested but not established on the small NHANES sub-panels, and is not observed on EHRShot. We did not test whether these tools had seen the external cohorts during pretraining, so we cannot separate a reranking benefit from any such exposure; the cohort-dependence of the effect is the more informative observation.

\subsection{A single-CBC-feature baseline on the development cohort}

The comparisons above rank candidate expressions against one another. To ground
them against the simplest marker a clinician could already use, we also scored
each CBC feature on its own. On the held-out development-cohort split we
computed the direction-agnostic univariate AUC of every individual CBC feature
and retained the best one per disease. The frontier LLM-tool candidates and the
scorer-ranked candidates were evaluated on that same split with the same metric,
so all columns of Table~\ref{tab:single_feature} are on a common scale. They
differ both in the pool over which each maximum is taken the 14 features, each tool's first five emitted candidates, and 25 scorer-selected candidates and in the rule that selects those candidates: emission order for the tool
columns, predicted score for the scorer column. Two of the tool cells are
\textit{n.a.} only because a tool's first five happened to be unevaluable on the
held-out split while the scorer's five, drawn from that same submitted list,
were evaluable. The selection-breadth analysis below applies to the
scorer-versus-single-feature gap only; it does not license reading the tool
columns as those tools' best achievable values.

The best single CBC feature proved a more demanding comparator than the
literature-seeded starting ratios used elsewhere in this work. It exceeded the
best of each tool's first five emitted candidates, across every evaluable tool,
in 8 of the 13 diseases and fell below it in the remaining 5, with no ties at full
precision. Averaged over
all 13 diseases, including the 5 in which it lost, the best single feature led
the best first-emitted tool candidate by $+1.80$\,\AUCpt{}; averaged over only the 8 diseases
in which it won, its margin was $+4.3$\,\AUCpt{}. The narrowest of those wins,
hyperthyroidism, is $+0.05$\,\AUCpt{} and is invisible at the three-decimal
precision printed in Table~\ref{tab:single_feature} (0.584 versus 0.584). An
unmodified CBC measurement is therefore already competitive with the expressions
these tools put forward first from pretrained knowledge alone. This comparator is
deliberately the first-emitted convention used in the reranking benchmark above,
not each tool's best of roughly 200 candidates, which would be higher.

The scorer-ranked candidate exceeded the best single feature in every disease,
by a mean of $+2.83$\,\AUCpt{}.No single feature was consistently strongest: the winning feature differed across diseases, most often RDW, MONOpct or HB.
Every column is selected on the split on which it is reported, so the absolute
values throughout Table~\ref{tab:single_feature} remain optimistic relative to a marker fixed in advance.

\begin{table}[!ht]
\centering
\scriptsize
\caption{Precision-at-K screening metrics for the discovered expressions, by
external cohort and disease.}
\label{tab:precision_at_k}
\begin{tabular}{llrrrrrr}
\toprule
Disease & Cohort & $n$ & $n_{\mathrm{pos}}$ & prec@0.5\% & prec@1\% & prec@5\% & enrich@1\% \\
\midrule
Hyperthyroidism & MIMIC-IV & 73,171 & 233 & 0.27\% & 0.82\% & 0.52\% & 2.57$\times$ \\
 & EHRShot & 3,308 & 51 & 0.00\% & 0.00\% & 1.82\% & 0.00$\times$ \\
\addlinespace
Type 1 diabetes & MIMIC-IV & 72,140 & 306 & 0.28\% & 0.14\% & 0.33\% & 0.33$\times$ \\
 & EHRShot & 3,186 & 88 & 0.00\% & 3.12\% & 3.77\% & 1.13$\times$ \\
\addlinespace
Familial Mediterranean fever & MIMIC-IV & 29,929 & 0 & \textit{n.a.} & \textit{n.a.} & \textit{n.a.} & \textit{n.a.} \\
 & EHRShot & 2,568 & 3 & 0.00\% & 0.00\% & 0.78\% & 0.00$\times$ \\
\addlinespace
Multiple sclerosis & MIMIC-IV & 72,712 & 126 & 0.27\% & 0.28\% & 0.22\% & 1.59$\times$ \\
 & EHRShot & 2,815 & 14 & 0.00\% & 0.00\% & 0.71\% & 0.00$\times$ \\
\addlinespace
Crohn's disease & MIMIC-IV & 71,091 & 243 & 0.28\% & 0.42\% & 0.45\% & 1.24$\times$ \\
 & EHRShot & 2,591 & 30 & 0.00\% & 0.00\% & 0.77\% & 0.00$\times$ \\
\addlinespace
Ulcerative colitis & MIMIC-IV & 69,635 & 210 & 0.57\% & 0.29\% & 0.32\% & 0.95$\times$ \\
 & EHRShot & 2,490 & 47 & 0.00\% & 4.00\% & 4.84\% & 2.12$\times$ \\
\addlinespace
Psoriasis & MIMIC-IV & 73,585 & 313 & 0.56\% & 0.56\% & 0.51\% & 1.32$\times$ \\
 & EHRShot & 2,859 & 69 & 0.00\% & 3.57\% & 3.60\% & 1.48$\times$ \\
 & NHANES & 16,997 & 464 & 4.71\% & 3.53\% & 3.65\% & 1.29$\times$ \\
\addlinespace
Rheumatoid arthritis & MIMIC-IV & 71,878 & 463 & 0.28\% & 0.42\% & 0.61\% & 0.65$\times$ \\
 & EHRShot & 2,769 & 93 & 0.00\% & 0.00\% & 2.90\% & 0.00$\times$ \\
 & NHANES & 38,887 & 2655 & 11.86\% & 12.08\% & 14.30\% & 1.77$\times$ \\
\addlinespace
Hashimoto thyroiditis & MIMIC-IV & 75,883 & 94 & 1.06\% & 0.79\% & 0.24\% & 6.38$\times$ \\
 & EHRShot & 3,327 & 24 & 0.00\% & 3.03\% & 0.60\% & 4.20$\times$ \\
\addlinespace
Celiac disease & MIMIC-IV & 70,115 & 100 & 0.00\% & 0.14\% & 0.11\% & 1.00$\times$ \\
 & EHRShot & 2,935 & 59 & 6.67\% & 3.45\% & 0.68\% & 1.72$\times$ \\
\addlinespace
Systemic lupus erythematosus & MIMIC-IV & 69,662 & 138 & 0.57\% & 0.72\% & 0.43\% & 3.62$\times$ \\
 & EHRShot & 2,635 & 118 & 0.00\% & 7.69\% & 11.36\% & 1.72$\times$ \\
\addlinespace
Systemic sclerosis & MIMIC-IV & 68,113 & 38 & 0.00\% & 0.00\% & 0.03\% & 0.00$\times$ \\
 & EHRShot & 2,708 & 20 & 0.00\% & 0.00\% & 1.48\% & 0.00$\times$ \\
\addlinespace
Sj\"ogren's syndrome & MIMIC-IV & 72,990 & 82 & 0.00\% & 0.00\% & 0.08\% & 0.00$\times$ \\
 & EHRShot & 2,901 & 36 & 0.00\% & 0.00\% & 1.38\% & 0.00$\times$ \\
\addlinespace
\bottomrule
\end{tabular}
\begin{flushleft}\footnotesize
\end{flushleft}
\end{table}

\begin{table}[!ht]
\centering
\small
\caption{Best single CBC feature (maximum over 14 features), the best candidate
from each frontier LLM tool (maximum over that tool's 5 candidates), and the
best scorer-ranked candidate (maximum over 25 candidates), all evaluated on the
same held-out development-cohort split.}
\label{tab:single_feature}
\resizebox{\textwidth}{!}{%
\begin{tabular}{lrrrrrr}
\toprule
 & Best single & GPT Deep & Gemini & SciSpace & Scorer- & $\Delta$ \AUCpt{} \\
Disease & CBC feature & Research & 3 Pro & BM Agent & ranked & (scorer $-$ single) \\
\midrule
Hyperthyroidism & 0.584 & 0.552 & 0.535 & 0.584 & 0.587 & $+0.25$ \\
Type 1 Diabetes & 0.594 & \textit{n.a.} & 0.576 & 0.597 & 0.609 & $+1.44$ \\
Familial Mediterranean Fever & 0.619 & 0.586 & 0.586 & 0.599 & 0.635 & $+1.65$ \\
Multiple Sclerosis & 0.560 & 0.584 & 0.576 & 0.576 & 0.587 & $+2.64$ \\
Crohn's Disease & 0.608 & 0.631 & 0.631 & 0.615 & 0.645 & $+3.69$ \\
Ulcerative Colitis & 0.642 & 0.664 & 0.664 & 0.664 & 0.685 & $+4.26$ \\
Psoriasis & 0.536 & 0.519 & 0.522 & 0.520 & 0.558 & $+2.23$ \\
Rheumatoid Arthritis & 0.615 & \textit{n.a.} & 0.599 & 0.588 & 0.643 & $+2.78$ \\
Hashimoto Thyroiditis & 0.612 & 0.543 & 0.550 & 0.544 & 0.632 & $+2.02$ \\
Celiac Disease & 0.663 & 0.588 & 0.570 & 0.570 & 0.671 & $+0.87$ \\
Systemic Lupus Erythematosus & 0.681 & 0.575 & 0.587 & 0.564 & 0.707 & $+2.55$ \\
Systemic Sclerosis & 0.698 & 0.735 & 0.640 & 0.677 & 0.754 & $+5.62$ \\
Sjögren's Syndrome & 0.656 & 0.545 & 0.545 & 0.595 & 0.723 & $+6.74$ \\
\bottomrule
\end{tabular}%
}
\end{table}

\subsection{Prediction-guided search improved post-hoc AUC}

\begin{figure}[!ht]
\centering
\includegraphics[width=\textwidth]{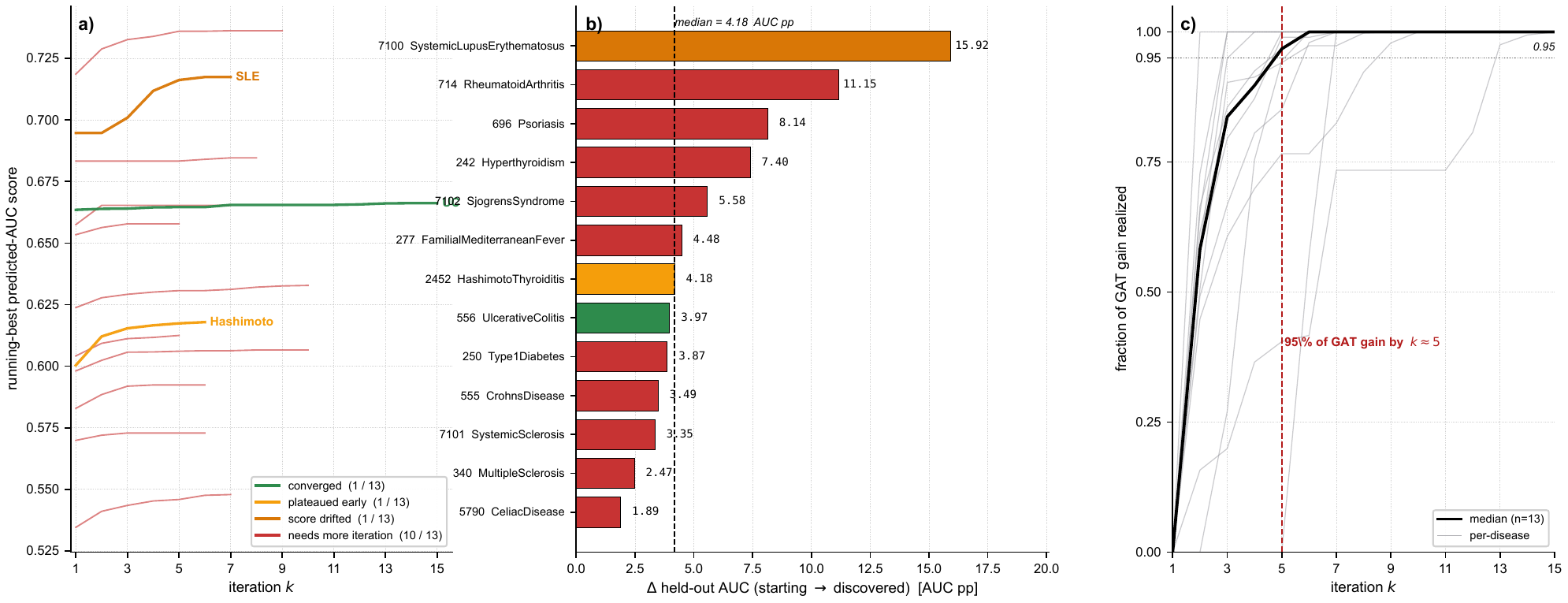}
\caption[Iteration trajectory and four-outcome audit summary.]{\textbf{Iteration
trajectory, AUC-pp lift, and plateau behavior across 13 immune-mediated diseases.}

\textbf{a)} Running-best predicted-AUC score over iterations $k$ for each
disease, coloured by a four-outcome audit summary; the legend labels correspond
to the audit classes of Table~\ref{tab:iter_outcomes} (converged =
\textsc{audit\_ok}, plateaued early = \textsc{terminate\_early}, score drifted =
\textsc{terminate\_late}, needs more iteration = \textsc{topup\_needed}).
\textbf{b)} Horizontal bar chart of $\Delta$ AUC from iteration~1 to the chosen iteration.
\textbf{c)} Fraction of total predicted-score gain realised by iteration $k$.
Thirteen thin grey lines, one per disease; the black line is the median across diseases.}

\label{fig:trajectory}
\end{figure}

The iterative search loop selected the best expression for each disease using only our GAT scorer's predicted AUC scores. After the loop terminated, the validation pipeline computed AUC for the initial expression and
for the discovered expression on the external validation datasets. Across the 13 diseases, the median gain in post-hoc realized AUC was $+4.18$\,\AUCpt{}, with a range from
$+1.89$\,\AUCpt{} for Celiac disease to
$+15.92$\,\AUCpt{} for SLE as described in Table~\ref{tab:iter_outcomes} and Figure~\ref{fig:trajectory}a,b. One of the 13, familial Mediterranean fever, is evaluable only on EHRShot and only on three positive cases, so its $+4.48$\,\AUCpt{} gain is not a stable estimate; it is reported for completeness and is marked as such throughout. Excluding it, the median across the remaining 12 diseases is $+4.08$\,\AUCpt{} and the mean is $+5.95$\,\AUCpt{}, so the headline does not depend on it. For SLE, the agent's literature-seeded starting expression ($\mathrm{NEUTpct}/\mathrm{LYMpct}$) had $0.516$ AUC and our tool-discovered expression $\frac{(\mathrm{RBC}+\mathrm{WBC})\cdot\mathrm{HB}}{\mathrm{RDW}+\mathrm{NEUTpct}+\mathrm{MONOpct}}$ had $0.676$ AUC; for celiac disease, the corresponding values were $0.538$ ($\mathrm{HB}/\mathrm{RDW}$) and $0.557$, with our tool-discovered expression $\frac{\mathrm{RBC}-\mathrm{RDW}}{\mathrm{RBC}+\mathrm{RDW}}$. 

Table~\ref{tab:discovered_main} lists, for each of the 13 diseases, the per-disease arithmetic expression selected by the iterative discovery loop, the candidate with the highest predicted-AUC score returned by the GAT scorer. Each loop starts from a simple, literature-motivated CBC ratio (the \emph{starting expression}) and iteratively proposes, scores, and refines candidate expressions over CBC features only. The \emph{GAT score} is the scorer's predicted-AUC for the discovered expression; the \emph{primary-cohort AUC} is the AUC of that same expression on the external validation (MIMIC-IV for all diseases except FMF, which is evaluated on EHRShot because of a lack of FMF cases in MIMIC), with 95\% bootstrap confidence intervals. AUC is the AUC value of the expression as a classifier for the disease. The 12 MIMIC-evaluable discovered expressions reproduce the corresponding values in Table~\ref{tab:iter_outcomes}; familial Mediterranean fever is shown as \textit{n.a.} here because its only evaluable cohort has three positive cases (see footnote).

\begin{table}[!ht]
\centering
\small
\caption{Per-disease outcomes of the model-guided search.}
\label{tab:iter_outcomes}
\begin{tabular}{llrrrrl}
\toprule
ICD & Disease & Iters & Starting expr & Discovered & $\Delta$ \AUCpt{} & Audit class \\
\midrule
242 & Hyperthyroidism & 10 & 0.527 & 0.602 & $+7.40$ & TOPUP\_NEEDED \\
250 & Type 1 Diabetes & 6 & 0.529 & 0.568 & $+3.87$ & TOPUP\_NEEDED \\
277 & Familial Mediterranean Fever & 5 & 0.684 & 0.729 & $+4.48$\textsuperscript{\textdagger} & TOPUP\_NEEDED \\
340 & Multiple Sclerosis & 6 & 0.545 & 0.570 & $+2.47$ & TOPUP\_NEEDED \\
555 & Crohn's Disease & 5 & 0.535 & 0.570 & $+3.49$ & TOPUP\_NEEDED \\
556 & Ulcerative Colitis & 15 & 0.532 & 0.572 & $+3.97$ & AUDIT\_OK  \\
696 & Psoriasis & 7 & 0.515 & 0.597 & $+8.14$ & TOPUP\_NEEDED \\
714 & Rheumatoid Arthritis & 10 & 0.503 & 0.615 & $+11.15$ & TOPUP\_NEEDED  \\
2452 & Hashimoto Thyroiditis & 6 & 0.564 & 0.606 & $+4.18$ & TERMINATE\_EARLY \\
5790 & Celiac Disease & 7 & 0.538 & 0.557 & $+1.89$ & TOPUP\_NEEDED \\
7100 & Systemic Lupus Erythematosus & 7 & 0.516 & 0.676 & $+15.92$ & TERMINATE\_LATE  \\
7101 & Systemic Sclerosis & 9 & 0.554 & 0.588 & $+3.35$ & TOPUP\_NEEDED \\
7102 & Sjögren's Syndrome & 8 & 0.541 & 0.596 & $+5.58$ & TOPUP\_NEEDED \\
\bottomrule
\end{tabular}
\end{table}

\begin{table}[ht]
\centering
\scriptsize
\caption{Discovered CBC biomarker expressions, top 1 per disease.}
\label{tab:discovered_main}
\resizebox{\textwidth}{!}{%
\begin{tabular}{llllrl}
\toprule
\textbf{ICD} & \textbf{Disease} & \textbf{Starting expression} & \textbf{Discovered expression} & \textbf{GAT score} & \textbf{Primary-cohort AUC [95\% CI]} \\
\midrule
242 & Hyperthyroidism & PLT*NEUTpct/LYMpct & (MCH*HB+HCT-RDW)*(HB+MCV*MCHC) & 0.607 & 0.602 [0.572, 0.636] (MIMIC-IV) \\
\addlinespace
250 & Type 1 diabetes & (PLT*NEUTpct)/LYMpct & (WBC+RDW+EOS\_pct)/(LYMpct+HB+HCT) & 0.592 & 0.568 [0.540, 0.596] (MIMIC-IV) \\
\addlinespace
277 & Familial Mediterranean fever & NEUTpct/LYMpct & RDW*NEUTpct*(RDW+RDW)/(MCV-RDW) & 0.613 & n.a.$^{\dagger}$ \\
\addlinespace
340 & Multiple sclerosis & NEUTpct/LYMpct & HB*RBC+LYMpct-NEUTpct & 0.573 & 0.570 [0.525, 0.610] (MIMIC-IV) \\
\addlinespace
555 & Crohn's disease & NEUTpct/LYMpct & (PLT*RDW+NEUTpct*RDW)/(LYMpct*HB+LYMpct) & 0.658 & 0.570 [0.532, 0.607] (MIMIC-IV) \\
\addlinespace
556 & Ulcerative colitis & NEUTpct/LYMpct & (PLT*RDW-LYMpct*HB)*(RDW*(RDW+RDW+NEUTpct))/(LYMpct*HB-MONOpct) & 0.666 & 0.572 [0.536, 0.604] (MIMIC-IV) \\
\addlinespace
696 & Psoriasis & MONOpct/LYMpct & (LYMpct+RDW+RDW)/(HB*HCT*MCV*NEUTpct*(BASO\_pct+EOS\_pct)) & 0.548 & 0.597 [0.566, 0.627] (MIMIC-IV) \\
\addlinespace
714 & Rheumatoid arthritis & (NEUT\_abs*PLT)/LYMpct & (RDW-RBC)*(MCV+RDW-MCH)/(HB*MCHC-RDW) & 0.633 & 0.615 [0.594, 0.635] (MIMIC-IV) \\
\addlinespace
2452 & Hashimoto thyroiditis & (PLT*NEUT\_abs)/LYMpct & (HB*HB*RBC)/(LYMpct*PLT+HCT) & 0.618 & 0.606 [0.556, 0.670] (MIMIC-IV) \\
\addlinespace
5790 & Celiac disease & HB/RDW & (RBC-RDW)/(RBC+RDW) & 0.665 & 0.557 [0.505, 0.609] (MIMIC-IV) \\
\addlinespace
7100 & Systemic lupus erythematosus & NEUTpct/LYMpct & ((RBC+WBC)*HB)/(RDW+NEUTpct+MONOpct) & 0.717 & 0.676 [0.633, 0.723] (MIMIC-IV) \\
\addlinespace
7101 & Systemic sclerosis & NEUTpct/LYMpct & ((RDW*MCV)-HB*MCH)/(HB*MCHC)*RDW & 0.736 & 0.588 [0.519, 0.658] (MIMIC-IV) \\
\addlinespace
7102 & Sjögren's syndrome & NEUTpct*PLT/LYMpct & RDW*MCV-HB*(HCT+RBC) & 0.685 & 0.596 [0.537, 0.652] (MIMIC-IV) \\
\bottomrule
\end{tabular}%
}
\begin{flushleft}\footnotesize
$^{\dagger}$\,Familial Mediterranean fever could only be evaluated on EHRShot, where it had just 3 positive cases. The resulting iterative-loop AUC was too unreliable (extremely wide confidence interval) to report, so it is given as \textit{n.a.}
\end{flushleft}
\end{table}

Because the discovered expression was selected from many candidates by predicted score alone, the size of the AUC gain depended on the starting expression. Diseases in which the starting CBC ratio was already informative, such as Hashimoto thyroiditis, left less room for absolute lift. Diseases in which the starting expression was near chance, such as RA and SLE, showed larger gains.

Most of the predicted-score improvement occurred early. Across diseases, a median of $k=5$ iterations captured 95\% of the per-disease predicted-score gain as shown in Figure~\ref{fig:trajectory}c. Twelve of 13 diseases reached this threshold by $k \leq 9$, and seven reached it by $k \leq 5$; Ulcerative Colitis (UC) was the lone late-saturating run, reaching the threshold at iteration 13. These trajectories suggest that a five-iteration budget captures most of the predicted-score signal.

\subsection{A random-scoring control isolates the scorer's contribution}
\label{subsec:random_control}

To test whether the iterative gains in Table~\ref{tab:iter_outcomes} come from the trained scorer rather than from the literature-grounded proposer or the iterative search, we reran the identical propose--score--refine loop on all 13 diseases with the GAT tool replaced by a random surrogate tool that draws each candidate's score uniformly from $[0.5,1.0]$. Holding the disease set, prompts,
feature allowlist, and external-validation pipeline fixed, the mean post-hoc gain across the 13 diseases collapsed from $+5.84$ to $+0.07$\,\AUCpt{} (median $+4.18$ to $-0.62$), and only 4 of 13 runs were nominally positive versus 13 of 13 for the GAT-guided search as detailed in Table~\ref{tab:iter_outcomes_random}.
These summaries include familial Mediterranean fever in both arms, for symmetry with the 13-disease presentation used for the headline gain; because that cell rests on three positive cases (Table~\ref{tab:iter_outcomes}), we also state the comparison without it. Excluding it from both arms leaves the contrast essentially unchanged: the mean collapses from $+5.95$ to $+0.13$\,\AUCpt{} (median $+4.08$ to $-0.51$), with 4 of 12 control runs nominally positive versus 12 of 12 GAT-guided, so the control conclusion does not depend on that cell.
The four positive deltas had bootstrap intervals overlapping their own starting expression's AUC and are consistent with proposer or seed variation rather than any scoring signal. Because random scoring rarely produces five consecutive no-improvement iterations, most control runs ran to the 15-iteration hard cap, whereas the GAT-guided runs in Table~\ref{tab:iter_outcomes} stopped well before the cap in
12 of 13 diseases. The control therefore behaved at chance, indicating that the gains in Table~\ref{tab:iter_outcomes} depend on the trained scorer rather than on the agentic scaffolding alone.

\begin{table}[!ht]
\centering
\small
\caption{Random-scoring control: per-disease outcomes when the trained GAT
scorer is replaced by a uniform random tool.}
\label{tab:iter_outcomes_random}
\begin{tabular}{llrrrr}
\toprule
ICD & Disease & Iters & Starting expr & Discovered & $\Delta$ \AUCpt{} \\
\midrule
242  & Hyperthyroidism              & 15 & 0.553 & 0.519 & $-3.41$ \\
250  & Type 1 Diabetes              & 15 & 0.531 & 0.525 & $-0.62$ \\
277  & Familial Mediterranean Fever & 15 & 0.684 & 0.677 & $-0.71$ \\
340  & Multiple Sclerosis           & 15 & 0.545 & 0.512 & $-3.30$ \\
555  & Crohn's Disease               & 15 & 0.535 & 0.552 & $+1.69$ \\
556  & Ulcerative Colitis           &  9 & 0.532 & 0.532 & $0.00$  \\
696  & Psoriasis                    & 15 & 0.510 & 0.576 & $+6.58$ \\
714  & Rheumatoid Arthritis         & 12 & 0.510 & 0.560 & $+4.98$ \\
2452 & Hashimoto Thyroiditis        & 15 & 0.584 & 0.543 & $-4.10$ \\
5790 & Celiac Disease               & 15 & 0.513 & 0.537 & $+2.45$ \\
7100 & Systemic Lupus Erythematosus & 15 & 0.517 & 0.513 & $-0.40$ \\
7101 & Systemic Sclerosis           & 15 & 0.554 & 0.545 & $-0.88$ \\
7102 & Sjögren's Syndrome            & 15 & 0.544 & 0.531 & $-1.38$ \\
\bottomrule
\end{tabular}
\begin{flushleft}\footnotesize
Columns are defined as in Table~\ref{tab:iter_outcomes}. The GAT scorer is replaced by a uniform random surrogate
drawing each candidate's score from $U[0.5,1.0]$; the disease set, prompts, feature allowlist, expression grammar,
batch-scoring interface and external-validation pipeline are unchanged. The starting expressions were re-derived
independently in this arm rather than copied from the main arm, and they differ from the main arm in 8 of the 13
diseases (242, 250, 696, 714, 2452, 5790, 7100, 7102), so the two arms are not paired at the starting point; this is
a limitation of the control, discussed in Methods.
\end{flushleft}
\end{table}

\subsection{Shapley attribution shows the discovered expressions largely recover known blood count biology}

\begin{figure}[p]
\centering
\includegraphics[height=0.85\textheight,keepaspectratio]{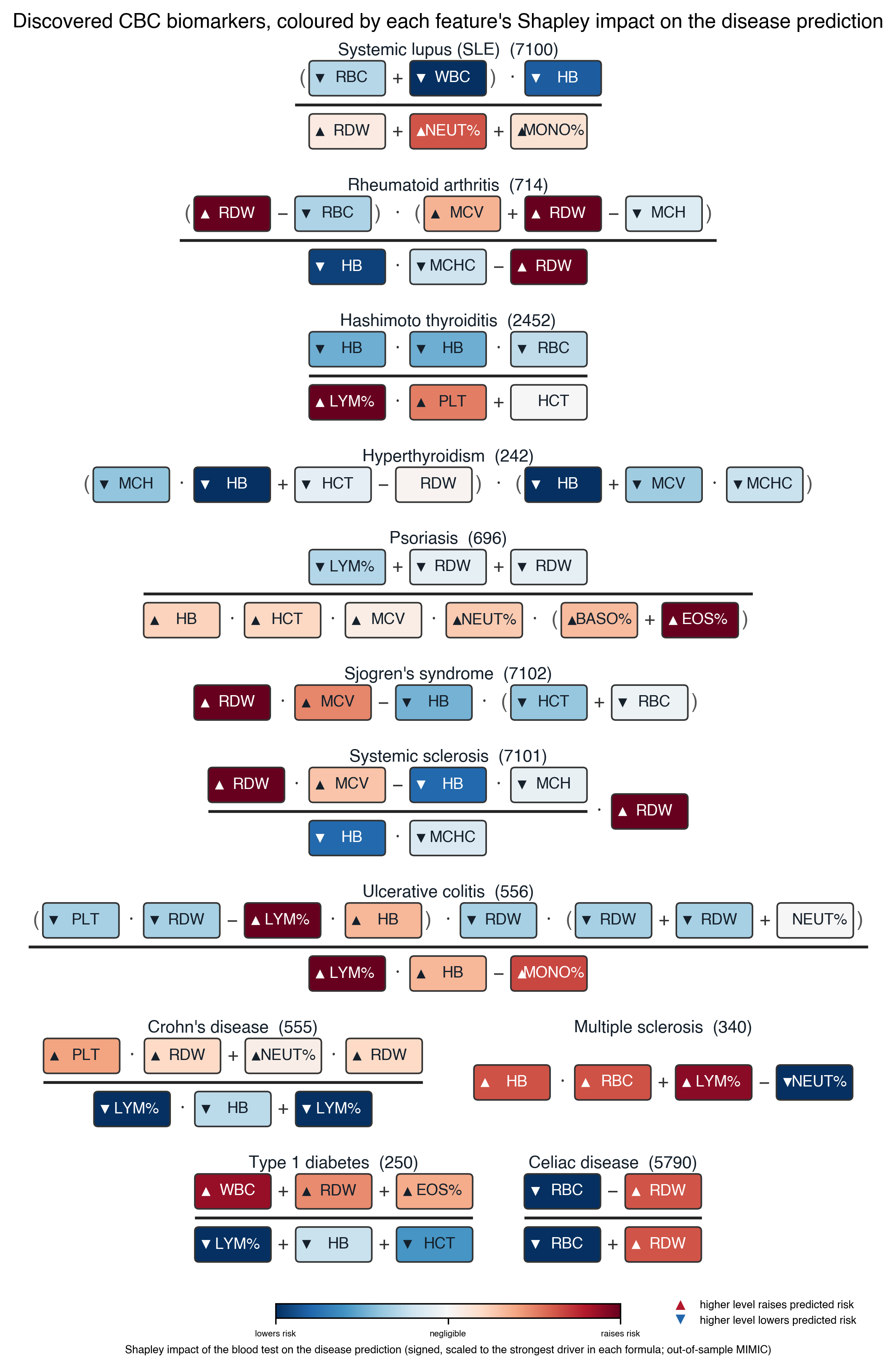}
\caption[Anatomy of the discovered CBC biomarkers.]{\textbf{Explanation of the discovered CBC biomarkers.} Each discovered expression is drawn with every blood-count leaf coloured by its signed Shapley contribution to the out-of-sample MIMIC-IV disease prediction (red, higher level raises predicted risk; blue, lowers it), scaled to the strongest driver within that expression; the triangle marks the direction.}
\label{fig:shapley_anatomy}
\end{figure}

The discovered expressions are put forward as \emph{interpretable}, but interpretability is only useful if the arithmetic recovers recognizable physiology rather than cohort-specific noise. We therefore asked what each blood-count feature contributes to a discovered expression's prediction and whether that contribution agrees with the clinical literature. We calculated the shapley values \cite{lundberg2017shap,strumbelj2014explaining} of the discovered expression. Because the attribution runs end to end through the expression, a feature's sign and magnitude reflect how it is actually used - numerator versus denominator, and the expression's overall orientation - rather than its marginal correlation with the label. Each signed contribution was then classified against the literature as \emph{Expected} (the model direction matches the established association), \emph{Surprising} (it contradicts the established direction, or the feature has no recognized link yet carries non-trivial weight), or \emph{Unclear}, through an intensive multi-source review with adversarial citation verification in which every supporting reference was re-resolved and unverifiable citations were discarded (Methods; Figure~\ref{fig:shapley_anatomy}, Table~\ref{tab:shapley_verdicts}).

Across the 64 feature contributions in the 12 expressions, 41 were expected, 21 Surprising, and 2 Unclear (Table~\ref{tab:shapley_verdicts}). The Expected majority was dominated by a single coherent axis, the anemia of inflammation: red-cell-distribution-width raised predicted risk and hemoglobin and red-cell terms lowered it, with leukocyte terms contributing in disease-specific directions (leukocytosis raising risk in most inflammatory diseases, leukopenia in lupus), recovering the hallmark hematological pattern of chronic immune-mediated disease. The two highest-AUC expressions were also the most biologically clean - every component of the systemic-lupus expression (AUC $0.676$) was expected, with low white-cell count (leukopenia) and anemia both raising predicted risk, and the rheumatoid-arthritis expression (AUC $0.615$) had five of six Expected components, anchored by red-cell-distribution-width elevation and hemoglobin reduction.

The Surprising minority was informative rather than reassuring. It clustered among several of the lower-AUC expressions - ulcerative colitis (five of six components Surprising), multiple sclerosis, and type~1 diabetes - where components carried non-trivial model weight in directions the literature does not support, consistent with overfitting to the discovery cohort. Concordance did not, however, track AUC monotonically: two of the lowest-AUC expressions, Crohn's disease and celiac disease, were themselves fully expected. Where a Surprising direction was clinically legible, it flagged a likely confounder rather than a novel mechanism: higher mean corpuscular volume raised predicted risk in Sj\"ogren's syndrome and systemic sclerosis (Table~\ref{tab:shapley_verdicts}), most plausibly reflecting treatment-related (for example, methotrexate) or age-related macrocytosis rather than a disease-intrinsic effect. Structurally, several expressions reused a feature across the fraction bar - for instance, red-cell-distribution-width appearing in both numerator and denominator - a pattern more consistent with symbolic-regression overfit than with added biology.

This analysis provides a fast, literature-grounded audit layer: it indicates that the highest AUC discovered expressions are consistent with recognized inflammatory-anemia physiology, and it localizes which components of the weaker expressions require scrutiny before any downstream use. Per-disease one-page explainers, with the full per-feature verdicts and verified citations, are provided as Supplementary Data~1.

\begin{table}[!ht]
\centering\footnotesize
\caption{Shapley attribution of each discovered biomarker's components and their consistency with the clinical literature.}
\label{tab:shapley_verdicts}
\begin{tabular}{llccl l}
\toprule
\textbf{Disease} & \textbf{AUC} & \textbf{Exp.} & \textbf{Surp.} & \textbf{Dominant driver} & \textbf{Notable surprising} \\
\midrule
Systemic lupus  & 0.676 & 6 & 0 & WBC~$\downarrow$ & --- \\
Rheumatoid arthritis  & 0.615 & 5 & 1 & RDW~$\uparrow$ & MCHC~$\downarrow$ \\
Hashimoto thyroiditis  & 0.606 & 3 & 2 & LYM\%~$\uparrow$ & LYM\%~$\uparrow$, HCT~$\uparrow$ \\
Hyperthyroidism  & 0.602 & 4 & 1 & HB~$\downarrow$ & RDW~$\uparrow$ \\
Psoriasis & 0.597 & 4 & 4 & EOS\%~$\uparrow$ & BASO\%~$\uparrow$, HB~$\uparrow$ \\
Sj\"ogren's syndrome  & 0.596 & 4 & 1 & RDW~$\uparrow$ & MCV~$\uparrow$ \\
Systemic sclerosis  & 0.588 & 4 & 1 & RDW~$\uparrow$ & MCV~$\uparrow$ \\
Ulcerative colitis  & 0.572 & 1 & 5 & LYM\%~$\uparrow$ & LYM\%~$\uparrow$, HB~$\uparrow$ \\
Crohn's disease  & 0.570 & 5 & 0 & LYM\%~$\downarrow$ & --- \\
Multiple sclerosis  & 0.570 & 0 & 3 & NEUT\%~$\downarrow$ & NEUT\%~$\downarrow$, HB~$\uparrow$ \\
Type 1 diabetes  & 0.568 & 3 & 3 & LYM\%~$\downarrow$ & WBC~$\uparrow$, RDW~$\uparrow$ \\
Celiac disease  & 0.557 & 2 & 0 & RBC~$\downarrow$ & --- \\
\midrule
\textbf{All 12 (64 components)} & --- & \textbf{41} & \textbf{21} & \multicolumn{2}{l}{2 Unclear} \\
\bottomrule
\end{tabular}
\end{table}

\section{Discussion}\label{sec:disc}

This study shows that a private clinical cohort can be effectively converted into a released, patient-level-free scoring resource for CBC-derived biomarker expressions. The GAT scorers were trained
inside the CHS data boundary from expression-level AUC labels computed on protected EHRs. After training, only the model weights are released, and even these weights do not hold sensitive information: the GAT tools were not trained on patient-level data, but on expression--AUC pairs calculated inside that boundary. External users or agents submit a disease identifier and a CBC expression and receive the predicted AUC score. The central contribution is an effective method for distilling private clinical evidence into a portable scorer that can rank agent-generated biomarker expressions without exposing the EHR data.

The clearest evidence for this release pattern comes from the external LLM-tool reranking benchmark. Across \nCells{llmtool_evaluable} verified-diagnosis comparisons, the GAT scorer selected an expression that outperformed the corresponding tool's expression in 67.9\%  [57.1\%,77.1\%] of cases, with a median paired gain of
$+1.50$\,\AUCpt{} (paired Wilcoxon $p = 4.0\times10^{-5}$). That advantage is carried by MIMIC-IV (86.1\% of \nCells{llmtool_mimic} comparisons) and is absent on EHRShot (48.7\% of \nCells{ehrshot_cells}, $p = 0.92$), so the benefit should be read as cohort-dependent rather than uniform. The EHRShot null result admits several non-exclusive explanations. First, statistical power there is limited: the EHRShot panels are roughly 25x smaller than their MIMIC-IV (mean 2,853 versus 71,748 patients per disease) and hold a median of only 47 positive cases per disease (range 3--118; Table~\ref{tab:cohort_sizes}), so paired AUC differences on EHRShot are estimated with wide uncertainty and a modest true reranking gain could go undetected.  This evaluation isolates the value of the private-cohort scorer from the generative capacity of the LLM tools: all candidate expressions came from external agentic systems, and the scorer only changed their ordering. The framework's robustness rests on the expressions using CBC values alone: no age, sex, or other demographic covariate enters either the scorer's training labels or the expressions it ranks. Excluding demographic covariates as inputs does not, however, establish the absence of demographic confounding; several of the red-cell indices used by the discovered expressions are themselves age-dependent, and this residual age confounding is discussed in Methods. 

The iterative discovery experiment provides a complementary demonstration of how such a scorer can guide agentic expression discovery. The search loop selected expressions using only predicted-AUC scores from the GAT. Post-hoc validation then compared the selected expression with the starting expression and found a median improvement of $+4.18$\,\AUCpt{} across the 13 disease runs ($+4.08$\,\AUCpt{} across the 12 runs remaining when the unstable familial Mediterranean fever cell is excluded; Table~\ref{tab:iter_outcomes}). We interpret this result as
selection-bound validation yield rather than autonomous clinical gain. The agent queried many candidate expressions, and the measured validation AUC was computed only after selection. The result, therefore, demonstrates that the scorer can guide search toward expressions with higher realized AUC in
this experimental setting, not that an unsupervised agent should be trusted to produce deployable biomarkers without audit.

A random tool negative control confirms that this guidance comes from the scorer rather than from the literature-grounded proposer or the iterative search: replacing the GAT with a uniform random tool while holding the loop, prompts, feature allowlist, and external-validation pipeline fixed collapsed the mean post-hoc gain across the 13 diseases from $+5.84$ to $+0.07$\,\AUCpt{}, and from $+5.95$ to $+0.13$\,\AUCpt{} when the unstable familial Mediterranean fever cell is excluded from both arms, the same exclusion applied to the headline median, with the surrogate score uncorrelated with realized AUC across tens of thousands of scored expressions and the few nominally positive control deltas having bootstrap intervals overlapping their own starting expression's AUC.

The magnitude of improvement also depended strongly on the starting point. Most starting expressions were the neutrophil-to-lymphocyte ratio or a closely related CBC ratio, a family of markers already known to carry
inflammatory signal. Diseases for which the starting expression was near chance, such as RA and SLE, left even more room for arithmetic improvement. Diseases with stronger starting ratios left less headroom. This pattern is consistent with the intended use of the scorer as a prioritization tool: it can help search the local
expression space around known CBC markers, but the size of the apparent gain depends on the baseline chosen for comparison.

The audit results reinforce the same interpretation, once the audit classes are read for what they measure. They describe search depth and a comparison against the independent LLM biomarker tools, not an improvement over each run's own starting expression: one of the 13 runs completed the full 15-iteration budget, one stopped early, one reached its predicted-score argmax before the blinded AUC trajectory had stopped improving, and the remaining ten would require further iterations before their outcome could be interpreted. On the comparison the gate actually makes, the result is uniformly negative: in all 10 runs where the audit located an independent-tool comparator set on the primary cohort, the discovered expression scored below the strongest comparator expression. This is a different and more demanding comparison than the reranking benchmark reported above, which asks only whether the scorer improves a tool's own ordering within that tool's candidate set; the audit gate asks whether a 15-iteration blinded search beats the best of roughly 200 expressions from three tools combined, and it does not. A third comparator, the best single CBC feature, is cleared on the development cohort: the scorer-ranked candidate exceeded it in every disease, by a mean of $+2.83$\,\AUCpt{}, of which roughly $+0.26$\,\AUCpt{} is attributable to its wider candidate pool under an independence null; hyperthyroidism is the one disease whose margin is not distinguishable from that selection artifact. This should not be read as a failure of the released scorer. Rather, it identifies the level at which the current system is ready to operate: ranking, triage, and hypothesis generation. Consistent with this readiness level, a Shapley attribution of the discovered expressions found that 41 of 64 feature contributions were concordant with the established clinical literature, while the Surprising minority clustered among several of the lower-AUC expressions and localised which components warrant audit (Figure~\ref{fig:shapley_anatomy}, Table~\ref{tab:shapley_verdicts}). The separation between predicted-score optimization and post-hoc realized-AUC evaluation is essential here: it allows the scorer to be used by an external agent while preserving an independent audit signal that can detect cases in which the predicted-score trajectory and realized validation performance diverge.

Several potential extensions could be explored in future work. First, the same release pattern could be implemented across multiple institutions to create an ensemble for improved scoring. Second, the expression primitives could be extended from the baseline CBC measurements to longitudinal CBC trajectories, which may better capture immune-mediated disease dynamics. Third, the vocabulary could be expanded
to other routinely measured panels, such as chemistry, urinalysis, or inflammatory markers, while retaining the same patient-row-free scoring interface. Fourth, the disease panel could be extended beyond the 13 conditions studied here. Finally, future work should train a single GAT for multiple diseases.

In conclusion, private EHR-derived evidence can be transformed into a publicly accessible biomarker discovery resource without revealing individual patient data, enabling public LLMs and agents to prioritize interpretable CBC expressions based on signals from a protected clinical cohort.

\section{Methods}\label{sec:methods}
\begingroup
\small
\setstretch{1.0}

\subsection{Study design and terminology}

This study evaluates whether a private clinical cohort can be converted into a released GAT-based scoring tool for CBC biomarker expressions. The scorer is a registry of disease-specific graph-attention networks trained inside the Clalit Health Services (CHS) data boundary. At inference, the scorer receives a disease identifier and a CBC biomarker candidate, represented as an arithmetic-expression tree, and returns the
predicted AUC score. It does not query patient rows, individual CBC measurements, or expression-level AUC labels.

We distinguish two AUC quantities throughout the study. First, the \emph{Clalit expression-label AUC} is the direction-agnostic univariate AUC used to supervise the disease-specific scorers during training.
Second, the \emph{external-evaluation AUC} is the AUC obtained when a fixed expression is evaluated without retraining in MIMIC-IV, EHRShot, or NHANES.

\subsection{Cohorts, phenotypes, and CBC measurements}

The private training source was the CHS ambulatory member panel of approximately 5{,}500{,}000 patients. For each disease, cases and controls were drawn within the CHS data boundary to form balanced 1:1 case-control subcohorts. The 13 immune-mediated phenotypes were indexed by International Classification of Diseases (ICD) code: Hyperthyroidism (242.0/.00/.01), Type~1 diabetes (250.x1/.x3, x=0--9), Familial Mediterranean fever (277.31), Multiple sclerosis (340), Crohn's disease (555.0/.1/.2/.9), Ulcerative colitis (556.0--556.6/.8/.9), Psoriasis (696.0/.1), Rheumatoid arthritis (714.0/.1/.2/.8/.81/.89/.9), Hashimoto thyroiditis (245.2), Celiac disease (579.0), Systemic lupus erythematosus (710.0; the lupus family also includes 695.4, discoid lupus erythematosus), Systemic sclerosis (710.1), and Sj\"ogren's syndrome (710.2). Per-disease CHS sample sizes are listed in Table~\ref{tab:cohort_sizes}.

The index date for a case was defined as the earliest diagnosis encounter, preceded by a 30-day blanking window to avoid CBC measurements concurrent with the diagnostic. The CBC input was the last CBC measurement before this index date within a five-year lookback window. Controls were sampled from patients without the target diagnosis, with index dates drawn from the empirical case index-date distribution to reduce calendar-time and seasonality bias. Case-control balancing was applied after cohort construction.

External evaluation used three cohorts. Each cohort was preprocessed separately for every disease, so the analysable $n$ varies across the panel and no single figure describes a cohort (Table~\ref{tab:cohort_sizes}). MIMIC-IV \cite{johnson2023mimic} contributed a mean of 71{,}748 critical-care patients per disease over the 12 evaluable diseases (median 72{,}009; range 68{,}113--75{,}883; the FMF panel of 29{,}929 patients holds no positive cases and is excluded), and EHRShot \cite{wornow2023ehrshot} a mean of 2{,}853 Stanford ambulatory patients per disease (median 2{,}815; range 2{,}490--3{,}327). The primary NHANES \cite{cdc1988nhanes} verified-diagnosis analysis was restricted to two cross-sectional phenotypes, Psoriasis ($n = 16{,}997$) and Rheumatoid arthritis ($n = 38{,}887$). Three further phenotypes: Hyperthyroidism (242), Type~1 diabetes (250), and Hashimoto thyroiditis (2452), do carry an NHANES label, but that label rests on a coarse self-report questionnaire item rather than a verified diagnosis, so they were excluded from the primary NHANES analysis and retained only as a sensitivity analysis (these items span $n = 49{,}510$ to $n = 52{,}231$ participants). For the remaining eight diseases (277, 340, 555, 556, 5790, 7100, 7101, 7102) no harmonized NHANES questionnaire item exists at all; these were reported as \textit{n.a.} for NHANES.

\subsection{CBC expression space}
Candidate biomarkers were represented as binary arithmetic-expression trees, a representation also used in symbolic regression and equation discovery \cite{cranmer2020pysr,udrescu2020feynman,petersen2021dsr}. Leaves were drawn from a 14-feature CBC vocabulary, and internal nodes were drawn from the four-operator alphabet $\{+,-,\times,\div\}$. Numeric literals were forbidden, so all candidates were constructed only from measured CBC features and arithmetic composition. The maximum tree depth was three. Expressions were canonicalized so that equivalent commutative forms were deduplicated.

The scorer is pretrained on a synthetic CBC-expression corpus of approximately two million depth-stratified expression trees per disease. Concretely, for each disease, 2{,}000{,}000 random synthetic expression trees were generated independently from these primitives. The synthetic expressions were split 80/20 into training and validation slices. The primitives were the 14-feature CBC lab panel (the manuscript's canonical CBC features: hemoglobin, hematocrit, white blood cell count, red blood cell count, platelets, MCV, MCH, MCHC, RDW, neutrophils, lymphocytes, monocytes, eosinophils, and basophils) combined with the four arithmetic operators ($+$, $-$, $\times$, $\div$). Trees are constructed at depths 1, 2, and 3 under the depth-3 cap that was used throughout the analyses; the universe of distinct depth-1, depth-2, and depth-3 trees over this vocabulary is 784, $\sim$2.55$\times$10$^6$, and $\sim$2.6$\times$10$^{13}$ respectively, so depth-3 must be subsampled. Stratified sampling preserves coverage of each depth band while keeping the per-disease training corpus computationally tractable; the depth fractions are approximately 5\% depth-1, 25\% depth-2, and 70\% depth-3 of the $\sim$2{,}000{,}000-tree corpus per disease; these fractions illustrate the depth-stratified sampling design as illustrated in Figure~\ref{fig:corpus_depth}. 

\begin{figure}[!ht]
\centering
\includegraphics[width=\textwidth]{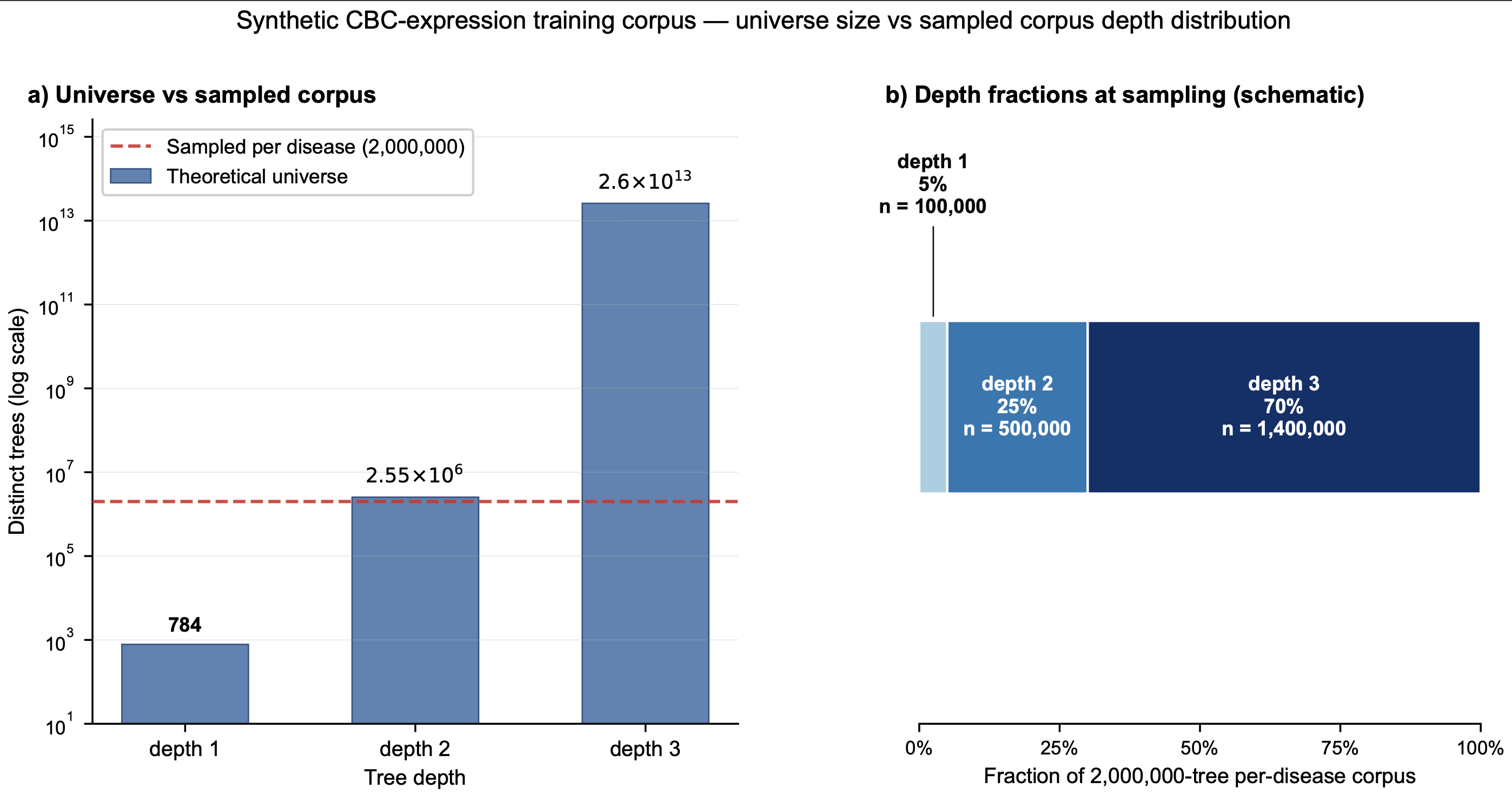}
\caption{Synthetic CBC expression training data. \textbf{a)}~Universe of distinct depth-stratified trees over the
14-feature CBC vocabulary $\times$ 4 arithmetic operators \textbf{b)}~Approximate depth fractions at sampling.}
\label{fig:corpus_depth}
\end{figure}

Within the CHS data boundary, each expression was evaluated on the corresponding balanced
case-control subcohort and labeled by AUC, defined as
\[
\mathrm{AUC}^{\pm}(e)=\max\{\mathrm{AUC}(e),1-\mathrm{AUC}(e)\}.
\]
This convention maps inverse and direct association to the same discovery score and constrains all target labels to the interval $[0.5,1.0]$.

\subsection{Released GAT scorer architecture and training}

Each disease-specific scorer was a graph-attention network (GAT)
\cite{velickovic2018gat,brody2022gatv2,li2022clinicalgat} operating on the parsed expression tree. Operator and CBC-feature primitives were represented as labeled graph nodes, with tree edges used to encode the expression structure.
The encoder used six graph-attention layers, eight attention heads, a 256-dimensional hidden state, and dropout of 0.1, followed by an attention-weighted global readout over all node representations. These parameters were chosen
by hyperparameter optimization to increase the correlation between predicted AUC and observed AUC. The scalar output was transformed to the interval $[0.5,1.0]$ to match the direction-agnostic AUC target.

Our GAT objective was designed to both score and rank the expressions produced by agents, combining pointwise regression with listwise ranking. For a training batch
\[
\mathcal{B}=\{(G_i,y_i)\}_{i=1}^{m},
\]
where $G_i$ is the parsed expression graph, $y_i \in [0.5,1.0]$ is the
direction-agnostic AUC target, and
\[
s_i=f_{\theta}(G_i)
\]
is the scorer prediction, the pointwise regression component was the Huber loss
\cite{huber1964}
\[
L_{\mathrm{Huber}}(\mathcal{B})
=
\frac{1}{m}\sum_{i=1}^{m}\ell_{\gamma}(s_i-y_i),
\]
with
\[
\ell_{\gamma}(r)=
\begin{cases}
\frac{1}{2}r^2, & |r|\leq \gamma,\\[4pt]
\gamma\left(|r|-\frac{1}{2}\gamma\right), & |r|>\gamma.
\end{cases}
\]
Here $r=s_i-y_i$ is the prediction residual and $\gamma$ is the Huber threshold.

The listwise ranking component was an $\alpha$-ListNet loss
\cite{cao2007listnet}. The target scores and predicted scores were converted into top-one probability distributions over the expressions in the list:
\[
p_i
=
\frac{\exp(\alpha y_i)}
{\sum_{j=1}^{m}\exp(\alpha y_j)},
\qquad
q_i
=
\frac{\exp(\alpha s_i)}
{\sum_{j=1}^{m}\exp(\alpha s_j)}.
\]
The ListNet loss was then the cross-entropy between the target-induced and prediction-induced distributions:
\[
L_{\mathrm{ListNet}}(\mathcal{B})
=
-\sum_{i=1}^{m} p_i \log q_i.
\]

The composite training objective was
\[
L(\mathcal{B})
=
\beta L_{\mathrm{Huber}}(\mathcal{B})
+
(1-\beta)L_{\mathrm{ListNet}}(\mathcal{B}),
\]
with $\beta=0.5$, and Huber threshold $\gamma=1$. The same architecture, loss weights, optimizer settings, learning-rate schedule, batch size, and epoch budget were used for all 13 diseases after hyperparameter optimization on the RA disease. Model selection used minimum validation loss on the held-out synthetic-expression slice, with rank correlation used as a
secondary tie-break.

After training, only the learned model weights and non-patient metadata were exported. Patient rows, individual CBC values, and expression-level CHS AUC labels remained inside the institutional environment.

\subsection{The tool scorer Application Programming Interface (API)}

The trained registry was exposed through a Model Context Protocol endpoint \cite{anthropic2024mcp}. The endpoint provided read-only metadata calls and scoring calls. Metadata calls enumerated the available
diseases, legal CBC features, operator alphabet, expression-depth bound, and checkpoint metadata. Scoring calls accepted one expression or a batch of expressions and returned predicted-AUC scores in $[0.5,1.0]$,
or a null value when an expression could not be parsed. The interface was therefore sufficient for ranking candidate CBC expressions but did not accept patient-level fields or return patient-derived records.

\subsection{LLM-guided propose--score--refine search}

The iterative discovery experiment used an agentic LLM search agent to propose CBC expressions and the released scorer to rank them.

Before proposing any expressions, the discovery agent performed a structured review of the public literature for each disease to assemble a disease-specific seed set of CBC ratios previously linked to inflammation or disease activity — such as, the neutrophil-to-lymphocyte, platelet-to-lymphocyte, and monocyte-to-lymphocyte ratios, red-cell-distribution-width indices, and composite inflammation scores. These literature-derived seeds initiated the first proposal round, anchoring the search in established hematological reasoning rather than arbitrary feature combinations; the review used only public sources and did not access the EHR data.

For each disease, the agent first received the legal expression grammar: the 14-feature CBC vocabulary, the four arithmetic operators, the maximum depth of three, and the prohibition on numeric literals. At each iteration, the agent proposed up to 200 candidate expressions. Invalid or duplicate expressions were removed, and the remaining candidates were scored in batches by the released GAT scorer. The automated validity filter enforced the feature allowlist, the operator set and the no-literals rule, but it did not enforce the depth bound, which was conveyed to the agent as part of the prompt rather than checked programmatically. Five of the 13 selected expressions (hyperthyroidism, ulcerative colitis, psoriasis, rheumatoid arthritis and systemic sclerosis) consequently exceed three operator levels, and two of those exceed the eight leaves a depth-three binary tree admits (ulcerative colitis, 11 leaves; psoriasis, 9). Because the scorer was trained only on depth-$\le 3$ trees, its predicted scores for those five expressions are extrapolations beyond the corpus grammar. We report this as a limitation of the search harness rather than retrospectively excluding the affected runs; the realized external AUCs in Table~\ref{tab:external_winner} are unaffected, since they are measured, not predicted. Structural extrapolation of this kind has two separable consequences for the scorer. Computationally, successfully parsed over-depth trees remain scoreable: the expression encoder operates on local bidirectional parent--child edges, and the scoring interface returns a range-bounded predicted AUC in $[0.5,1.0]$, so an out-of-grammar expression receives an ordinary-looking score rather than an explicit out-of-distribution warning. Predictively, however, that score is unvalidated. The scorers were trained only on depth-$\le 3$ binary trees, whereas the five affected expressions have depth four or five; the ulcerative-colitis expression contains 11 leaves and 21 nodes, compared with at most eight leaves and 15 nodes under the training grammar. The concern is not connectivity, the encoder's six message-passing rounds span even the deepest selected tree, and the attention-weighted graph-level readout aggregates all node representations but distribution shift: every attention weight and readout statistic was fitted exclusively to graphs of at most 15 nodes, and graph-neural-network accuracy is documented to degrade silently on graphs larger than those seen in training \cite{yehudai2021size,xu2021extrapolate}. Descriptively, among the twelve evaluable runs in Table~\ref{tab:discovered_main}, the five depth-$>3$ cases include both the largest predicted-minus-external-AUC difference (systemic sclerosis, $+14.8$\,\AUCpt{}) and the only negative difference (psoriasis, $-4.9$\,\AUCpt{}). This pattern is suggestive of instability, but it cannot establish increased variance or absence of bias, because it comprises only five selected expressions and conflates scorer error with cross-cohort transfer. A production deployment of the harness or any agentic hook built on it should therefore enforce the depth bound in the same programmatic validation layer that enforces the feature allowlist, the operator set, and the no-literals rule, rejecting or explicitly flagging out-of-grammar candidates before scoring.

The prompt history for the next iteration contained only predicted-AUC scores from the released scorer. Specifically, it included the top expressions observed so far by predicted score and a small set of recent
unique candidates. It did not include CHS expression-label AUCs, external-evaluation AUCs, bootstrap intervals, case counts, or any patient-level information. The loop was configured to stop when the best predicted-AUC score had nominally plateaued after five consecutive iterations without strict improvement, or when it reached a configurable hard cap of 15 iterations, whichever came first. In practice the agent applied the plateau criterion loosely, so the realised patience varied between runs: 11 of the 13 runs terminated between zero and three iterations after the highest-scoring candidate had been found, one (celiac disease) matched the nominal five-iteration patience, and one (ulcerative colitis) ran to the 15-iteration cap. Run lengths therefore ranged from 5 to 15 iterations, and the Iters column of Table~\ref{tab:iter_outcomes} should be read as what each run actually did rather than as the output of a strictly enforced stopping rule. The discovered expression for a disease was the valid expression with the highest predicted-AUC score observed during the run.

After the loop terminated, the starting expression and the discovered expression were evaluated to obtain the external-evaluation AUCs. These AUCs were computed only after candidate selection and were not visible to the agent during proposal, ranking, or early stopping.

\subsection{Post-hoc audit of iterative search runs}

Each iterative search run was assigned a post-hoc audit class from two quantities recorded after the loop terminated: the number of distinct iterations the run completed, and a strict finalize gate. The finalize gate compared the discovered expression's external-evaluation AUC on the run's primary cohort against the strongest comparator available for that disease and cohort, defined as the highest external-evaluation AUC attained by any expression in the frozen candidate sets of the three independent LLM biomarker tools. The gate is a strict comparison of point estimates and does not use confidence intervals; where the audit harness did not locate a comparator file for the primary cohort, it returned a pass by default. Runs were labelled \textsc{audit\_ok} when the run completed the full 15-iteration budget, \textsc{terminate\_early} when it stopped short of that budget but passed the finalize gate, and \textsc{topup\_needed} otherwise, meaning the run neither exhausted the budget nor cleared the comparator. A \textsc{terminate\_early} run was subsequently relabelled \textsc{terminate\_late} when the external-evaluation AUC at the run's final iteration exceeded that of the highest-predicted-score expression by more than the half-width of a paired bootstrap interval excluding zero, indicating that the predicted-score argmax was reached before the agent-blinded AUC trajectory stopped improving. The audit class is therefore a statement about search depth and about a comparison with independent LLM tools; it is not a statement that a run improved on its own starting expression, and it should not be read as one. On the comparison the gate does make, the outcome was uniformly negative: in all 10 of the 13 runs where the audit located a comparator set on the primary cohort, the discovered expression scored below the strongest comparator expression, and the three runs recorded as passing the gate did so only by the default described above. These classes were used only for post-hoc interpretation and did not affect candidate generation.

\subsection{Random-scoring ablation}
\label{subsec:methods_random_control}

To attribute the iterative gains to the trained scorer rather than to the iterative deep research, we ran a negative-control arm in which the GAT scorer was replaced by a uniform random scorer that draws each candidate's score independently from $U[0.5,1.0]$. Everything else was held fixed the same agentic propose--score--refine skill, disease set, frozen task prompts, 14-feature CBC allowlist, depth-3 no-literals grammar, batch-scoring interface, and external-validation pipeline so that the per-expression score source was the only deliberate difference. The starting expression was not fixed across the two arms. Because the literature-review step that seeds iteration~1 is itself agent-generated and not deterministic, the control arm proposed a different starting ratio in 8 of the 13 diseases (242, 250, 696, 714, 2452, 5790, 7100, 7102); comparing Table~\ref{tab:iter_outcomes} with Table~\ref{tab:iter_outcomes_random} shows the corresponding starting-expression AUCs differ in those eight runs and agree in the remaining five (277, 340, 555, 556, 7101). Every reported $\Delta$ is computed within its own arm, against that arm's own starting expression, so the comparison of gains remains paired within each run. As in the main runs, realized external-validation AUC was logged post hoc but never exposed to the proposer. We used the skill with the same arguments (five iterations before early stopping, or a 15-iteration hard cap), because random scores rarely produce five-in-a-row strict no-improvement. Most control runs reached the 15-iteration cap rather than early stopping. Over tens of thousands of scored expressions, the random score was effectively uncorrelated with realized AUC.

\subsection{External validation of discovered expressions}

The starting expression and discovered expression from each disease run were evaluated without retraining in MIMIC-IV, EHRShot, and NHANES when the corresponding disease label was available and had sufficient positive case counts. External datasets were normalized to the same CBC feature convention used for Clalit. Expressions were evaluated by a restricted arithmetic interpreter using the same direction-agnostic AUC convention as in training.

For each evaluable disease--cohort cell, we report the external evaluation AUC for the starting expression and the discovered expression, together with percentile bootstrap 95\% confidence intervals using 500 resamples.
Comparisons with unavailable labels, undefined expression values, or insufficient positive case counts were reported as \textit{n.a.}. NHANES main-text external validation was restricted to the verified-diagnosis phenotypes psoriasis and Rheumatoid arthritis.

\subsection{External LLM-tool reranking benchmark}

The reranking benchmark tested whether the released scorer could improve the ordering of candidate expressions generated by independent LLM biomarker tools. Three external tools were used: GPT Deep Research, Gemini 3 Pro, and SciSpace BM Agent. Each tool was asked to generate a set of legal CBC arithmetic expressions for each disease using a frozen task prompt. The prompt was optimized before the benchmark was frozen using the GAAPO procedure \cite{secheresse2025gaapo}. This benchmark did not use the iterative search agent and did not involve propose--score--refine feedback.

For each tool produced candidate set, the GAT scorer selected the expression with the highest predicted-AUC score. We then compared two expressions from the same candidate set: the tool's first-emitted expression and the scorer-selected expression. Both expressions were evaluated on the corresponding external cohort using the same external validation pipeline. The paired effect size was:
\[
\Delta_{\mathrm{tool}} =
\mathrm{AUC}^{\pm}_{\mathrm{external}}(\mathrm{scorer\ pick})
-
\mathrm{AUC}^{\pm}_{\mathrm{external}}(\mathrm{tool\ first\ pick}).
\]
Thus, the released scorer affected only which expression was selected. All reported differences were based on realized external-evaluation AUC.

\paragraph{Benchmark cell accounting.} The benchmark grid is \nCells{external_cohorts} external cohorts $\times$ \nCells{disease_panel} diseases $\times$ \nCells{external_llm_tools} tools $= 117$ candidate cells. NHANES provides a harmonized outcome label for only 5 of the 13 diseases (242, 250, 696, 714, 2452), so the \nCells{nhanes_unlabeled_icds} unlabelled diseases remove $8\times3=24$ cells and leave \nCells{cartesian_total} cells with data. MIMIC-IV contains no positive familial Mediterranean fever (ICD~277) cases, so the direction-agnostic univariate AUC is undefined there and the three MIMIC-IV $\times$ FMF cells are unevaluable, leaving 90. Of those 90, nine (hyperthyroidism 242, type~1 diabetes 250, and Hashimoto thyroiditis 2452, each $\times$ \nCells{external_llm_tools} tools, all on NHANES) rest on a coarse self-report questionnaire item rather than a verified diagnosis. Consistent with the NHANES restriction applied throughout this work, these nine cells are excluded from the primary panel and reported separately as a sensitivity sub-panel. The primary reranking panel is therefore \nCells{llmtool_evaluable} cells $=$ \nCells{llmtool_mimic} MIMIC-IV $+$ \nCells{ehrshot_cells} EHRShot $+$ \nCells{llmtool_nhanes_primary} NHANES verified-diagnosis, and the sensitivity panel is \nCells{llmtool_nhanes_sensitivity} cells. Paired Wilcoxon signed-rank tests are computed within a panel; the primary and sensitivity panels are never pooled into a single test, and the combined \nCells{llmtool_combined_reference}-cell figure is quoted for reference only.

\subsection{Single-CBC-feature baseline}

Every column compared in Table~\ref{tab:single_feature} uses the same metric on
the same data: the direction-agnostic univariate AUC, $\max(\mathrm{AUC},\,1 -
\mathrm{AUC})$, of a single predictor on the held-out development-cohort split,
the convention applied to expressions throughout this work. The columns differ in the candidate pool over which the maximum is taken and in the rule selecting the candidates: emission order for the tool columns,
predicted score for the scorer column. (i) The best single-feature baseline is the maximum over the 14 individual CBC features; no model was fitted, no demographic covariate was included and no feature was
transformed, so this baseline measures what an unmodified CBC measurement achieves on its own. (ii) Each frontier-tool column is the maximum over that tool's first five emitted candidates for that disease the same first-emitted convention as the reranking benchmark, and not that tool's best of roughly 200. (iii) The scorer-ranked column is the maximum over 25 candidates: the scorer's five highest-scoring picks from each of five candidate generators, namely the three frontier tools' full candidate lists, every expression each tool submitted is scored, and the scorer keeps five, and two internal language-model generators, a base generator and a
literature-enhanced variant. The scorer-ranked pool is therefore wider than, and not contained in, the tool
pools. The generator supplying the scorer-ranked maximum was the internal literature-enhanced generator in 12 of the 13 diseases and the SciSpace BM Agent list in hyperthyroidism.

Because each column is a maximum selected on the split on which it is reported, all values are optimistic relative to a marker fixed in advance, and the optimism grows with pool size. To estimate the part of the scorer-versus-single-feature gap attributable to pool size alone, we simulated the null in which no candidate carries signal: for each disease, we drew direction-agnostic null AUCs as $0.5 + |Z|\sigma$, $Z \sim \mathcal{N}(0,1)$, with $\sigma = \sqrt{(n_{+}+n_{-}+1)/(12\,n_{+}n_{-})}$ the standard error of the AUC statistic
at that disease's realized 1:1 held-out split size ($n = 597$ to $28{,}089$),
and recorded the difference between the maximum of 25 draws and the maximum of
14 draws over 20{,}000 replicates (seed 20260429). Draws were independent,
whereas real candidates within a disease share features and are correlated, so the resulting figure is a reference for the selection component rather than an exact correction.

\subsection{Shapley attribution and literature-concordance protocol}

To interpret the discovered expressions, we attributed each expression's disease prediction to its constituent CBC features and tested each attribution against the clinical literature. 

For each of the 12 MIMIC-IV-labeled diseases, the discovered expression $E$ was evaluated for each patient, standardized, and its orientation was fixed so that higher values indicate higher disease probability, consistent with the direction-agnostic AUC convention. We then fit a one-predictor logistic disease model, $P(\mathrm{disease})=\sigma(\beta_0+\beta_1 z)$, where $z$ is the standardized, orientation-fixed expression value; the single predictor is the whole expression, not the individual features. This prediction was attributed back to the \emph{raw} CBC features appearing in $E$ using exact interventional Shapley values \cite{lundberg2017shap,strumbelj2014explaining}, with the cohort median as the reference patient. Because each expression uses at most eight distinct features ($d\le 8$), the Shapley sum over all $2^{d}$ feature coalitions was computed exactly rather than approximated. Attributing through the composite $E$, rather than to each feature marginally, means a feature's signed contribution reflects its placement in the numerator or denominator and the expression's orientation as actually used. For visualization (Figure~\ref{fig:shapley_anatomy}), the signed contributions within each expression were scaled to that expression's strongest driver, mapping to $[-1,1]$ (red, raises predicted risk; blue, lowers it).

Each signed contribution was classified against the clinical literature as \emph{Expected} (the model direction matches the established direction of that blood-count feature in that disease), \emph{Surprising} (it contradicts the established direction, or the feature has no recognized association yet carries non-trivial model weight), or \emph{Unclear} (evidence too thin or conflicting). Classifications were produced by an intensive multi-agent literature review over PubMed and web sources. Language agents can synthesize scientific literature at a level comparable to subject-matter experts, but general-purpose models fabricate a large fraction of the citations they emit unless generation is grounded in retrieved sources, and even retrieval-grounded agents recover the relevant literature only partially \cite{loke2026can,asai2026openscholar,skarlinski2024paperqa2}. The review was therefore followed by an adversarial verification pass in which every cited reference was re-resolved by identifier and any citation that could not be confirmed to support the stated direction was discarded; per-feature verdicts and the surviving verified citations are reported in the per-disease one-page explainers - Supplementary Data~1.

This attribution is an interpretive audit layer and carries the corresponding caveats: the logistic link is fit in-sample on the MIMIC-IV cohort it explains; the expressions were discovered on Clalit but attributed on MIMIC-IV; age confounds several red-cell indices; and a feature recurring in both numerator and denominator may be an overfit artifact rather than an independent signal. The reported concordance is therefore evidence that an expression recovers recognized physiology, not evidence of causation. A further caveat concerns the acuity of the attributing population. MIMIC-IV samples critical-care admissions, so the CBC values entering the attribution are drawn during acute illness, and the patients carrying an ICD code for a chronic autoimmune disease are acutely ill inpatients, typically older, more comorbid, and more heavily medicated than the ambulatory prevalent cases of the Clalit cohort on which the expressions were discovered. Feature contributions estimated in this setting can therefore load on correlates of acuity, treatment exposure, and age structure rather than on primary disease physiology. The MCV$\uparrow$ contributions flagged as Surprising in Sj\"ogren's syndrome and systemic sclerosis (Table~\ref{tab:shapley_verdicts}) illustrate this concretely: macrocytosis is a recognized consequence of ageing and of antifolate or thiopurine immunosuppressants such as methotrexate and azathioprine that are commonly used in these diseases, so a positive MCV contribution estimated among critical-care admissions is at least as consistent with medication or age effects as with disease-intrinsic biology, as flagged in Results. Because the Expected/Surprising classification is made against literature on primary disease biology, it cannot by itself separate a genuinely novel association from such acuity-linked confounding; re-estimating the attributions on an ambulatory cohort would disambiguate the two, but the ambulatory EHR cohort available to this study, EHRShot, holds a median of only 47 positive cases per disease (Table~\ref{tab:cohort_sizes}), too few for stable per-feature attribution across the panel, so ambulatory re-attribution remains an open validation step. Familial Mediterranean fever was excluded because MIMIC-IV contains no positive cases.

\subsection{Privacy analysis}\label{ssec:privacy}

The released scorers were never trained on patient-level data, and the exported artifact contains only trained weights and non-patient metadata; the public interface returns predicted AUC scores for submitted expression trees. The training input for each disease is a corpus of approximately two million candidate expressions, each labeled by a single cohort-level summary statistic: the direction-agnostic case--control AUC of that expression, computed inside the CHS boundary. Trained models can in principle leak information about their training data~\cite{shokri2017mia,carlini2021extracting}, but here the training data itself contains nothing patient-specific to leak: even an adversary who recovered every training pair verbatim would obtain arithmetic expressions and their cohort-level AUCs the same class of aggregate performance summary that clinical prediction studies, including this one, publish routinely and no patient's laboratory values, diagnoses, dates, or identity. Nor can any single patient leave a meaningful imprint on the labels: each label is an AUC over the disease's balanced case--control subcohort (2{,}390--111{,}835 patients; Table~\ref{tab:cohort_sizes}), so replacing one patient's values shifts any label by no more than about one part in a thousand many times smaller, in every disease, than the released scorer's own approximation error (per-disease mean absolute error $0.0030$--$0.0186$; Table~\ref{tab:training_diag}).

What the weights can reveal is therefore what they were built to reveal: which CBC expressions separate cases from controls in each disease. Cohort-composition quantities cannot be recovered beyond what is already public, the case--control ratio was fixed at 1:1 by design, disease prevalence in the source population never enters the training pipeline, and the subcohort sizes are disclosed in Table~\ref{tab:cohort_sizes}. An empirical membership-inference or extraction audit would accordingly probe an attack surface, a model trained on individual patient records that this design does not create.

\section*{Data availability}
MIMIC-IV \cite{johnson2023mimic} is available to PhysioNet-credentialed researchers at \url{https://physionet.org/content/mimiciv/}.  \\
EHRShot \cite{wornow2023ehrshot} is available through the Stanford Medicine Data Use Agreement; cohort construction follows the EHRShot benchmark
protocol. NHANES \cite{cdc1988nhanes} is public domain via the U.S.\
Centers for Disease Control and Prevention at \url{https://wwwn.cdc.gov/nchs/nhanes/}; due to national and organizational data privacy regulations, CHS individual-level data from this study cannot be shared publicly.

\section*{Code availability}
The scoring tool and the trained models that support this study are openly available as a self-contained software and model release at \url{https://github.com/SeffiCohen/gat_agent_tool}. The release comprises (i) the \texttt{gat-agent-tool} Python package, which exposes the trained graph-attention scorer to any
large-language-model agent through a Python API, an MCP server, and HuggingFace and OpenAI tool-calling drivers; (ii) the model-definition and expression-to-graph encoding code required to load and run the scorer; (iii) the thirteen disease-specific trained graph-attention checkpoints reported here, one per phenotype; and (iv) the two agent skills used in this study \\
\texttt{biomarker-discovery}, which runs the propose--score--refine search loop against the released scorer, and \\ 
\texttt{biomarker-explainer}, which produces the Shapley attribution and literature-concordance analysis of Figure~\ref{fig:shapley_anatomy} and Table~\ref{tab:shapley_verdicts}. 
Consistent with the privacy design of the method, only the trained weights and the scoring code
are released: no patient-level records, individual complete-blood-count values, or expression-level CHS AUC labels are included, and scoring a candidate expression requires only a single forward pass over its symbolic graph encoding.

\section*{Acknowledgements}
This research was supported by The Israel Science Foundation (grant No. 2672/24).

\section*{Author contributions}

{\sloppy
S.C.: Conceptualization, Methodology, Software, Formal analysis, Investigation,
Validation, Visualization, Writing -- original draft, Writing -- review \&
editing, Project administration. L.A.F.: Data curation, Methodology Review A.A.: Data curation, methodology, Review. R.J.: Methodology, review. M.M.L.: Methodology, review, Data curation. A.N.:   Methodology, review. B.R.:  Methodology, review, supervision. R.B.: Methodology, review, supervision, resources.
N.D.: Conceptualization, Data curation, Resources, Methodology, Supervision, Writing -- review \& editing. M.Z.: Conceptualization, Methodology,
Supervision, Writing -- review \& editing. 
N.D.\ and M.Z.\ contributed equally as supervising authors.
\par}

\section*{Competing interests}
The authors declare no patents or patent applications relating to the released
scorer, the trained checkpoints, or the biomarker expressions reported here.
Apart from the affiliations stated above, the authors declare no competing
interests.

\section*{Ethics declarations}

The Clalit Health Services (CHS) component of this study was approved by the Clalit Community Clinics Helsinki committee under approval number 1COM0161-19. The work is a retrospective analysis of de-identified electronic health record data held and analyzed entirely within the CHS data boundary; no participant was contacted, and no prospective or interventional procedure was performed. All CHS analyses were conducted in accordance with the Declaration of Helsinki and with applicable Israeli national and CHS institutional data-protection regulations.

\begingroup

\let\OLDthebibliography\thebibliography
\renewcommand{\thebibliography}[1]{\OLDthebibliography{#1}\setlength{\parskip}{1pt}\setlength{\itemsep}{1pt}}
\bibliographystyle{naturemag}
\bibliography{main}
\endgroup

\end{document}


\maketitle
\thispagestyle{fancy}
Each section below is a self-contained explainer for one discovered expression: the
biomarker drawn as a glass-box tree whose leaves are coloured by each blood test's Shapley
contribution to the disease prediction, an Expected / Surprising / Unclear classification of
every component against the clinical literature, and the supporting evidence. Shapley values are
computed out-of-sample on MIMIC-IV.

Every citation was resolved against PubMed by identifier and independently re-verified; the bibliography at the end is shared across all twelve explainers.

\footnote{\textbf{The reports generated by the skill \texttt{biomarker-explainer} are provided "as is," along with the references, and may contain errors.}}

\clearpage
\section*{Systemic lupus (SLE)}
\addcontentsline{toc}{section}{Systemic lupus (SLE) (7100)}
\begin{center}
\includegraphics[width=0.75\textwidth]{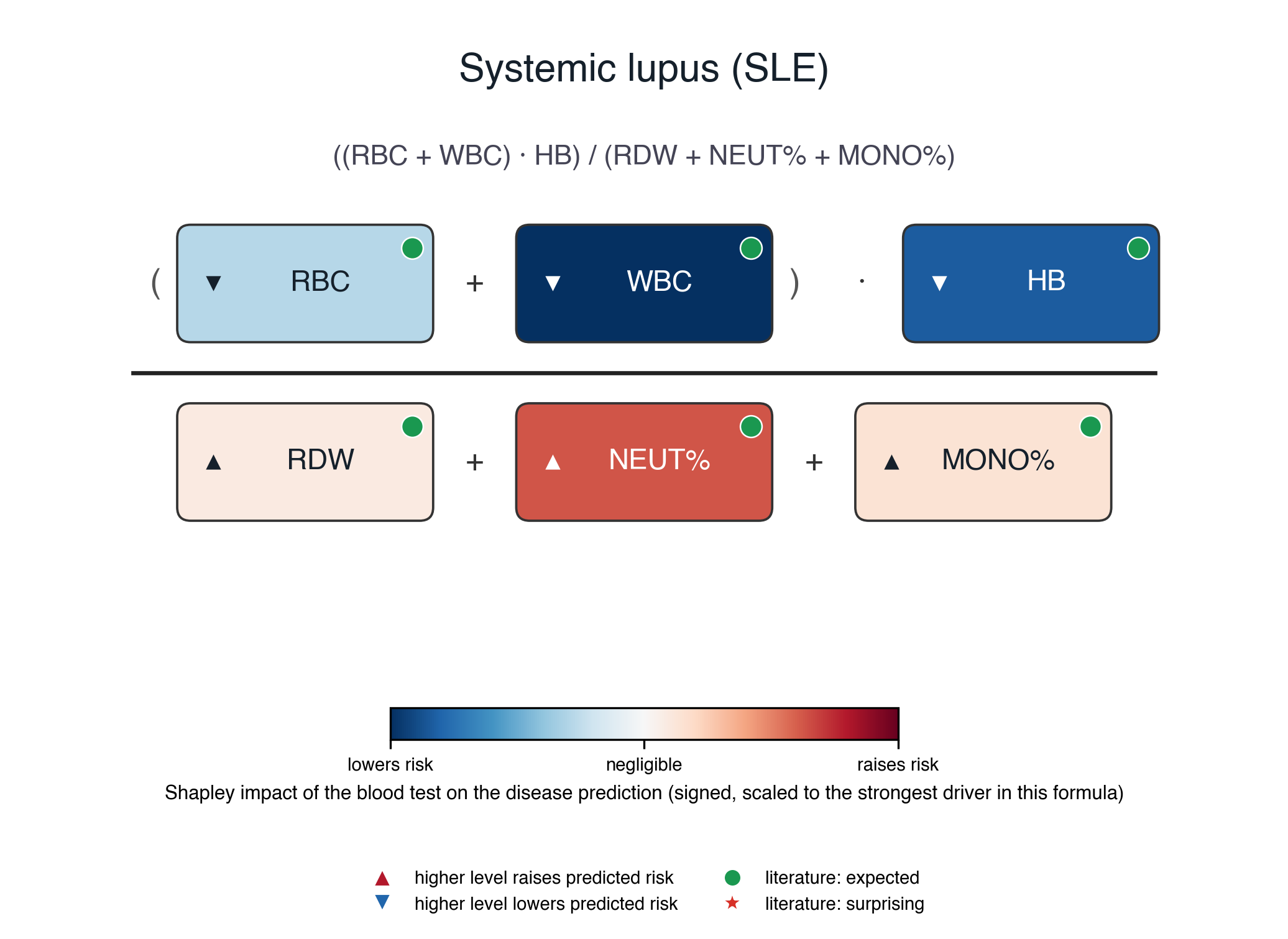}
\end{center}
{\footnotesize Each blood-test leaf is coloured by its Shapley impact on the disease prediction (red = raises risk, blue = lowers it; triangle = direction; circle = expected, star = surprising).\par}
\vspace{2pt}
\textbf{Biomarker:} ((RBC + WBC) \ensuremath{\cdot} HB) / (RDW + NEUT\% + MONO\%) $\cdot$ out-of-sample AUC (MIMIC) $\approx$ 0.6757\par
\textbf{WBC} (numerator). Higher WBC lowers the predicted risk --- \textbf{expected}, consistent with WBC being reduced in Systemic lupus (SLE) (high confidence). Leukopenia is a core, criterion-level hematologic feature of SLE. \cite{carli2015leukopenia}\par
\textbf{HB} (numerator). Higher HB lowers the predicted risk --- \textbf{expected}, consistent with HB being reduced in Systemic lupus (SLE) (high confidence). In SLE, hemoglobin is characteristically reduced. \cite{kisaoglu2022hematologic}\par
\textbf{NEUT\%} (denominator). Higher NEUT\% raises the predicted risk --- \textbf{expected}, consistent with NEUT\% being elevated in Systemic lupus (SLE) (high confidence). In SLE the neutrophil fraction of the differential tends to rise for two converging reasons. (1) The lymphocyte denominator falls: T/B lymphopenia is one of the most consistent hematologic features of active SLE (immune-complex-mediated lymphocyte destruction, anti-lymphocyte antibodies, type I interferon-driven lymphocyte redistribution/apoptosis), which mechanically raises the neutrophil percentage. (2) The neutrophil compartment is activated and frequently expanded in active disease: SLE patients show elevated absolute neutrophil counts associated with neutrophil-activation markers (serum calprotectin), enrichment for low-density granulocytes, increased NETosis, and type I IFN activity. \cite{han2020neutrophil}\par
\textbf{RBC} (numerator). Higher RBC lowers the predicted risk --- \textbf{expected}, consistent with RBC being reduced in Systemic lupus (SLE) (high confidence). Anemia (reduced RBC count and hemoglobin) is one of the most common laboratory abnormalities in SLE, present in roughly 50\% of patients. \cite{voulgarelis2000anaemia}\par
\textbf{MONO\%} (denominator). Higher MONO\% raises the predicted risk --- \textbf{expected}, consistent with MONO\% being elevated in Systemic lupus (SLE) (medium confidence). In SLE the relative monocyte burden is increased and rises with disease activity, driven by type-I-interferon-mediated monocyte activation and expansion of pro-inflammatory CD14+CD16+ intermediate/non-classical and layilin+ monocyte subsets, plus monocyte recruitment into target organs (notably the kidney in lupus nephritis). \cite{mercadersalvans2023blood}\par
\textbf{RDW} (denominator). Higher RDW raises the predicted risk --- \textbf{expected}, consistent with RDW being elevated in Systemic lupus (SLE) (high confidence). In SLE, chronic systemic inflammation (elevated CRP, ESR, IL-6 and other cytokines) suppresses erythropoiesis and blunts the erythropoietin response, impairs iron mobilization (functional iron deficiency / anemia of chronic disease), and shortens red-cell survival. \cite{mercadersalvans2024red}\par
\vspace{2pt}
\clearpage
\section*{Rheumatoid arthritis}
\addcontentsline{toc}{section}{Rheumatoid arthritis (714)}
\begin{center}
\includegraphics[width=0.88\textwidth]{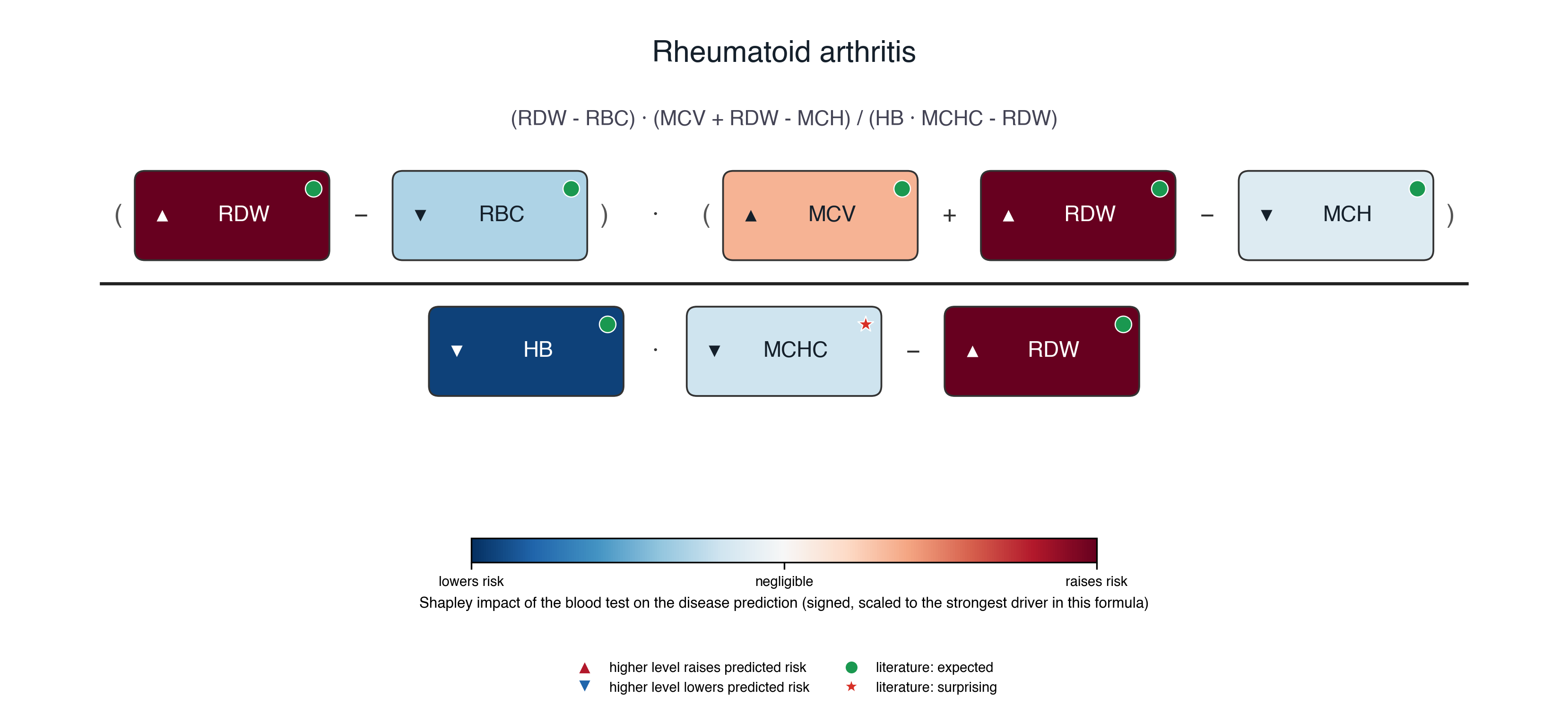}
\end{center}
{\footnotesize Each blood-test leaf is coloured by its Shapley impact on the disease prediction (red = raises risk, blue = lowers it; triangle = direction; circle = expected, star = surprising).\par}
\vspace{2pt}
\textbf{Biomarker:} (RDW - RBC) \ensuremath{\cdot} (MCV + RDW - MCH) / (HB \ensuremath{\cdot} MCHC - RDW) $\cdot$ out-of-sample AUC (MIMIC) $\approx$ 0.6146\par
\textbf{RDW} (numerator+denominator). Higher RDW raises the predicted risk --- \textbf{expected}, consistent with RDW being elevated in Rheumatoid arthritis (high confidence). RA is a chronic systemic inflammatory disease in which sustained pro-inflammatory cytokine activity (IL-6, TNF-alpha) suppresses effective erythropoiesis and disrupts iron metabolism (functional iron deficiency / anemia of chronic disease), producing a more heterogeneous circulating red-cell population. \cite{alrawi2018red}\par
\textbf{HB} (denominator). Higher HB lowers the predicted risk --- \textbf{expected}, consistent with HB being reduced in Rheumatoid arthritis (high confidence). Hemoglobin is lowered in rheumatoid arthritis primarily through anemia of chronic disease (ACD). \cite{wilson2004prevalence}\par
\textbf{MCV} (numerator). Higher MCV raises the predicted risk --- \textbf{expected}, consistent with MCV being variably altered in Rheumatoid arthritis (high confidence). Two competing mechanisms. \cite{vreugdenhil1990anaemia}\par
\textbf{RBC} (numerator). Higher RBC lowers the predicted risk --- \textbf{expected}, consistent with RBC being reduced in Rheumatoid arthritis (high confidence). Rheumatoid arthritis is a chronic systemic inflammatory disease, and anemia (low hemoglobin and reduced red cell count) is its single most common extra-articular/hematologic manifestation, classically of the anemia-of-chronic-disease (ACD) type. \cite{wilson2004prevalence}\par
\textbf{MCHC} (denominator). Higher MCHC lowers the predicted risk --- \textbf{surprising}, since MCHC is typically variably altered in Rheumatoid arthritis (high confidence). MCHC is not an RA-specific analyte. \cite{wilson2004prevalence}\par
\textbf{MCH} (numerator). Higher MCH lowers the predicted risk --- \textbf{expected}, consistent with MCH being reduced in Rheumatoid arthritis (medium confidence). Active RA drives a chronic IL-6/TNF-alpha-mediated inflammatory state that upregulates hepatic hepcidin. \cite{jeffrey1953some}\par
\vspace{2pt}

\clearpage
\section*{Hashimoto thyroiditis}
\addcontentsline{toc}{section}{Hashimoto thyroiditis (2452)}
\begin{center}
\includegraphics[width=0.62\textwidth]{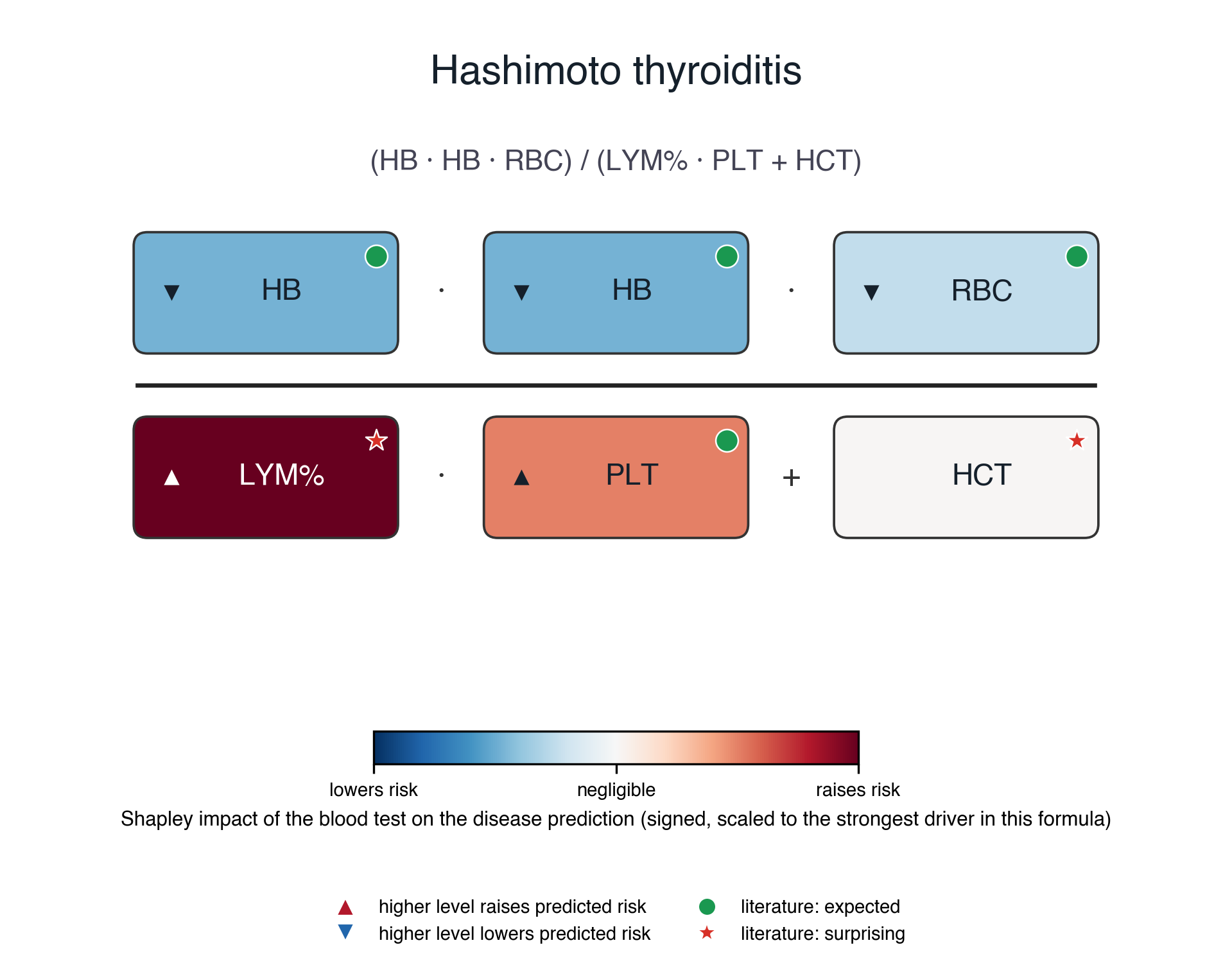}
\end{center}
{\footnotesize Each blood-test leaf is coloured by its Shapley impact on the disease prediction (red = raises risk, blue = lowers it; triangle = direction; circle = expected, star = surprising).\par}
\vspace{2pt}
\textbf{Biomarker:} (HB \ensuremath{\cdot} HB \ensuremath{\cdot} RBC) / (LYM\% \ensuremath{\cdot} PLT + HCT) $\cdot$ out-of-sample AUC (MIMIC) $\approx$ 0.6056\par
\textbf{LYM\%} (denominator). Higher LYM\% raises the predicted risk --- \textbf{surprising}, since LYM\% is typically reduced in Hashimoto thyroiditis (high confidence). Hashimoto thyroiditis (HT) is a chronic autoimmune thyroiditis in which autoreactive T and B lymphocytes home from the circulation into the thyroid gland to form the characteristic dense lymphocytic infiltrate. \cite{xue2025lymphocyte}\par
\textbf{PLT} (denominator). Higher PLT raises the predicted risk --- \textbf{expected}, consistent with PLT being elevated in Hashimoto thyroiditis (medium confidence). Hashimoto thyroiditis is a chronic systemic-inflammatory autoimmune state. \cite{cao2022platelet}\par
\textbf{HB} (numerator). Higher HB lowers the predicted risk --- \textbf{expected}, consistent with HB being reduced in Hashimoto thyroiditis (high confidence). In Hashimoto thyroiditis, hemoglobin tends to be lower / anemia is more common than in euthyroid controls. \cite{wopereis2018relation}\par
\textbf{RBC} (numerator). Higher RBC lowers the predicted risk --- \textbf{expected}, consistent with RBC being reduced in Hashimoto thyroiditis (high confidence). Thyroid hormones (T3/T4) drive erythropoiesis both directly --- stimulating proliferation of erythroid precursors and increasing cellular oxygen demand --- and indirectly by enhancing renal erythropoietin (EPO) production. \cite{bremner2012significant}\par
\textbf{HCT} (denominator). Higher HCT raises the predicted risk --- \textbf{surprising}, since HCT is typically reduced in Hashimoto thyroiditis (high confidence). Hashimoto thyroiditis is the leading cause of (subclinical/overt) hypothyroidism. \cite{wopereis2018relation}\par
\vspace{2pt}

\clearpage
\section*{Hyperthyroidism}
\addcontentsline{toc}{section}{Hyperthyroidism (242)}
\begin{center}
\includegraphics[width=1.00\textwidth]{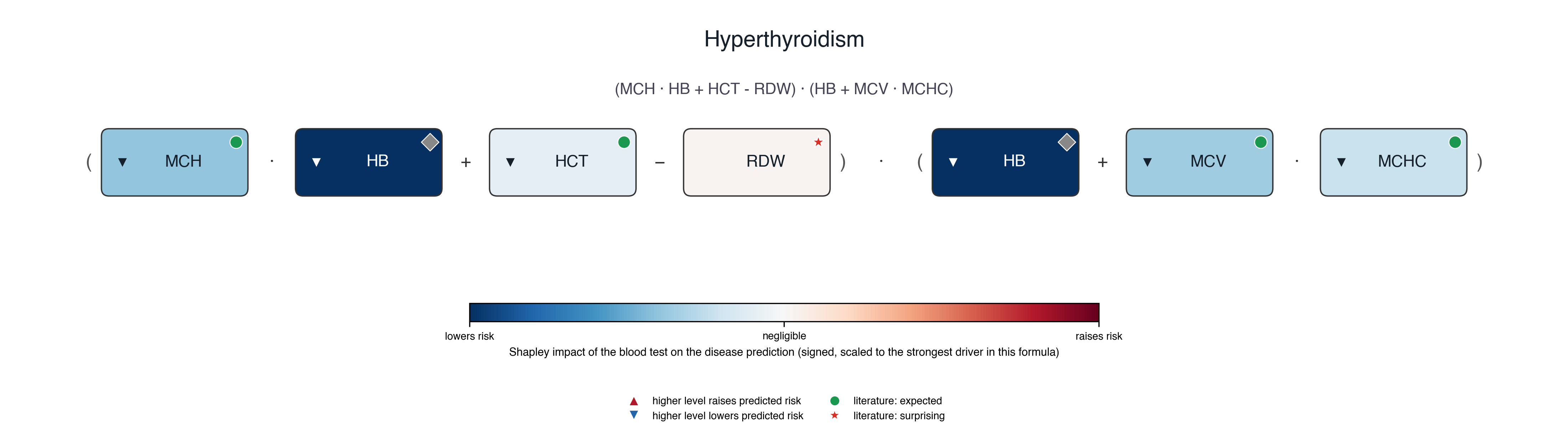}
\end{center}
{\footnotesize Each blood-test leaf is coloured by its Shapley impact on the disease prediction (red = raises risk, blue = lowers it; triangle = direction; circle = expected, star = surprising).\par}
\vspace{2pt}
\textbf{Biomarker:} (MCH \ensuremath{\cdot} HB + HCT - RDW) \ensuremath{\cdot} (HB + MCV \ensuremath{\cdot} MCHC) $\cdot$ out-of-sample AUC (MIMIC) $\approx$ 0.6015\par
\textbf{HB} (numerator). Higher HB lowers the predicted risk --- \textbf{unclear}, as the literature on HB in Hyperthyroidism is limited (medium confidence). Thyrotoxicosis perturbs erythropoiesis and iron handling in ways that, on balance, tend to LOWER hemoglobin and raise anemia prevalence relative to euthyroid controls. \cite{wopereis2018relation}\par
\textbf{MCH} (numerator). Higher MCH lowers the predicted risk --- \textbf{expected}, consistent with MCH being reduced in Hyperthyroidism (medium confidence). Hyperthyroidism accelerates erythropoiesis and red-cell turnover and is associated with relative/absolute iron deficiency and altered iron handling (thyroid hormone influences iron metabolism, hepcidin and erythropoietin signaling), producing iron-restricted, hypochromic/microcytic erythropoiesis. \cite{omar2010erythrocyte}\par
\textbf{MCV} (numerator). Higher MCV lowers the predicted risk --- \textbf{expected}, consistent with MCV being reduced in Hyperthyroidism (high confidence). In thyrotoxicosis, accelerated erythropoiesis and increased red-cell turnover driven by excess thyroid hormone, together with functional/relative iron deficiency (elevated demand, altered iron handling) and thyroid-hormone-induced bone-marrow hematopoietic stem/progenitor cell cycle arrest, produce smaller red cells. \cite{how1979red}\par
\textbf{MCHC} (numerator). Higher MCHC lowers the predicted risk --- \textbf{expected}, consistent with MCHC being reduced in Hyperthyroidism (low confidence). Thyroid hormones regulate erythropoiesis. \cite{yang2025impact}\par
\textbf{HCT} (numerator). Higher HCT lowers the predicted risk --- \textbf{expected}, consistent with HCT being variably altered in Hyperthyroidism (high confidence). HCT in hyperthyroidism is governed by two opposing forces. (1) Erythropoiesis-stimulating arm: thyroid hormone directly stimulates erythroid precursors (beta2-adrenergic mediated) and raises erythropoietin, so thyrotoxicosis can produce mild erythrocytosis / increased red-cell mass with erythroid marrow hyperplasia (Das 1975; Sullivan \& McDonald 1992) --- this would RAISE HCT. (2) Anemia-promoting arm that dominates at the population level: overt hyperthyroidism is consistently associated with higher anemia prevalence and lower hemoglobin, driven by plasma-volume expansion (hemodilution lowering measured HCT despite a normal/raised red-cell mass), altered iron metabolism, increased oxidative stress, accelerated red-cell turnover, and ineffective erythropoiesis when hematinic nutrients (iron, B12, folate) are deficient. \cite{das1975erythropoiesis}\par
\textbf{RDW} (numerator). Higher RDW raises the predicted risk --- \textbf{surprising}, since RDW is typically elevated in Hyperthyroidism (high confidence). RDW (anisocytosis) rises mainly with HYPOthyroidism and autoimmune thyroiditis, driven by impaired/ineffective erythropoiesis, reduced erythropoietin drive, and concomitant iron/B12/folate deficiency and autoimmune comorbidity, producing a heterogeneous RBC population. \cite{szczepanekparulska2017anemia}\par
\vspace{2pt}
\clearpage
\section*{Psoriasis}
\addcontentsline{toc}{section}{Psoriasis (696)}
\begin{center}
\includegraphics[width=1.00\textwidth]{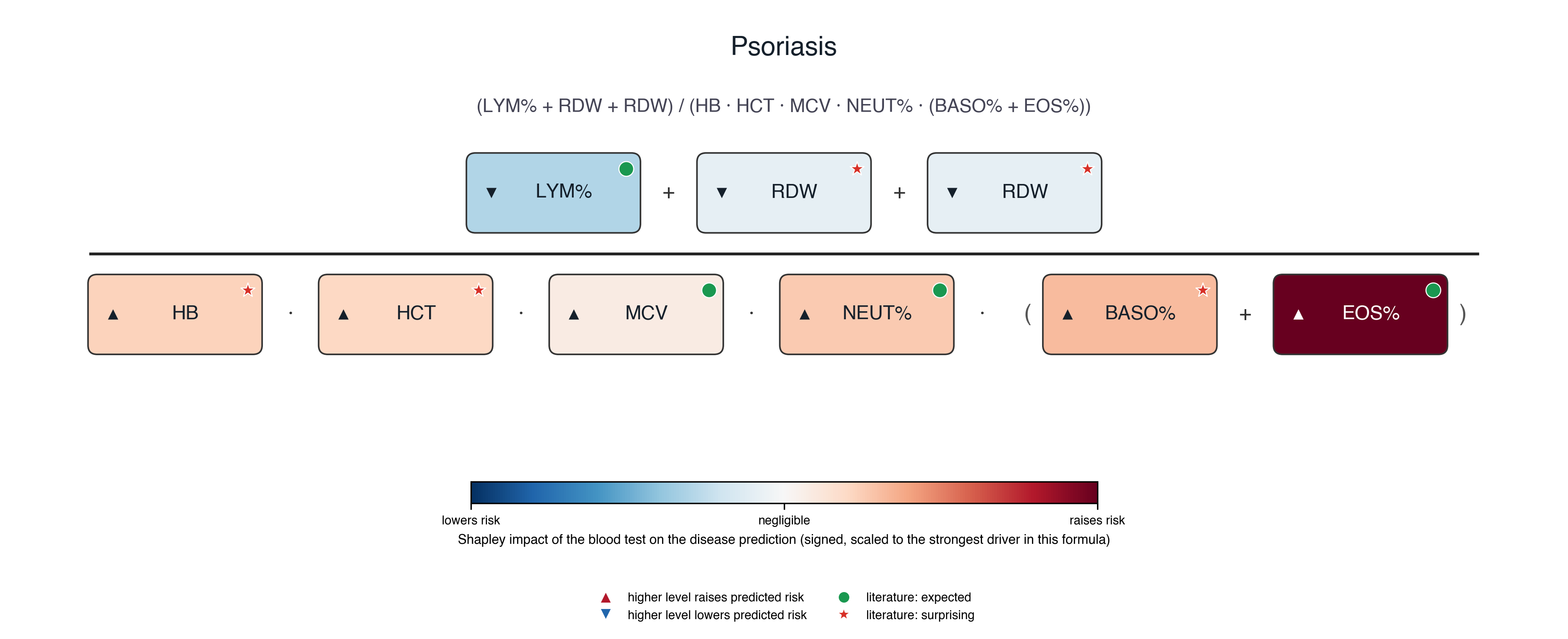}
\end{center}
{\footnotesize Each blood-test leaf is coloured by its Shapley impact on the disease prediction (red = raises risk, blue = lowers it; triangle = direction; circle = expected, star = surprising).\par}
\vspace{2pt}
\textbf{Biomarker:} (LYM\% + RDW + RDW) / (HB \ensuremath{\cdot} HCT \ensuremath{\cdot} MCV \ensuremath{\cdot} NEUT\% \ensuremath{\cdot} (BASO\% + EOS\%)) $\cdot$ out-of-sample AUC (MIMIC) $\approx$ 0.5966\par
\textbf{EOS\%} (denominator). Higher EOS\% raises the predicted risk --- \textbf{expected}, consistent with EOS\% being elevated in Psoriasis (high confidence). Psoriasis is a chronic immune-mediated inflammatory disease. \cite{zhou2023exploring}\par
\textbf{BASO\%} (denominator). Higher BASO\% raises the predicted risk --- \textbf{surprising}, since BASO\% is typically reduced in Psoriasis (high confidence). Psoriasis is a Th1/Th17-driven autoimmune disease (IL-23/IL-17 axis, neutrophil- and monocyte-dominated infiltrate), NOT a Th2/IgE/atopic process. \cite{arango2024biometrics}\par
\textbf{LYM\%} (numerator). Higher LYM\% lowers the predicted risk --- \textbf{expected}, consistent with LYM\% being reduced in Psoriasis (high confidence). Psoriasis is a chronic Th17/Th1-driven systemic inflammatory disease. \cite{wang2021neutrophil}\par
\textbf{NEUT\%} (denominator). Higher NEUT\% raises the predicted risk --- \textbf{expected}, consistent with NEUT\% being elevated in Psoriasis (high confidence). Psoriasis is driven by the IL-23/IL-17 inflammatory axis. \cite{liu2024associations}\par
\textbf{HB} (denominator). Higher HB raises the predicted risk --- \textbf{surprising}, since HB is typically reduced in Psoriasis (high confidence). Psoriasis is a chronic Th1/Th17-driven systemic inflammatory disease. \cite{lee2021low}\par
\textbf{HCT} (denominator). Higher HCT raises the predicted risk --- \textbf{surprising}, since HCT is typically reduced in Psoriasis (high confidence). Psoriasis is a chronic systemic inflammatory disease. \cite{lee2021low}\par
\textbf{RDW} (numerator). Higher RDW lowers the predicted risk --- \textbf{surprising}, since RDW is typically elevated in Psoriasis (high confidence). Psoriasis is a chronic systemic inflammatory disease. \cite{yi2022comparison}\par
\textbf{MCV} (denominator). Higher MCV raises the predicted risk --- \textbf{expected}, consistent with MCV being elevated in Psoriasis (medium confidence). In psoriasis, MCV tends to be modestly elevated relative to healthy controls. \cite{demirpektas2018evaluation}\par
\vspace{2pt}

\clearpage
\section*{Sjogren's syndrome}
\addcontentsline{toc}{section}{Sjogren's syndrome (7102)}
\begin{center}
\includegraphics[width=1.00\textwidth]{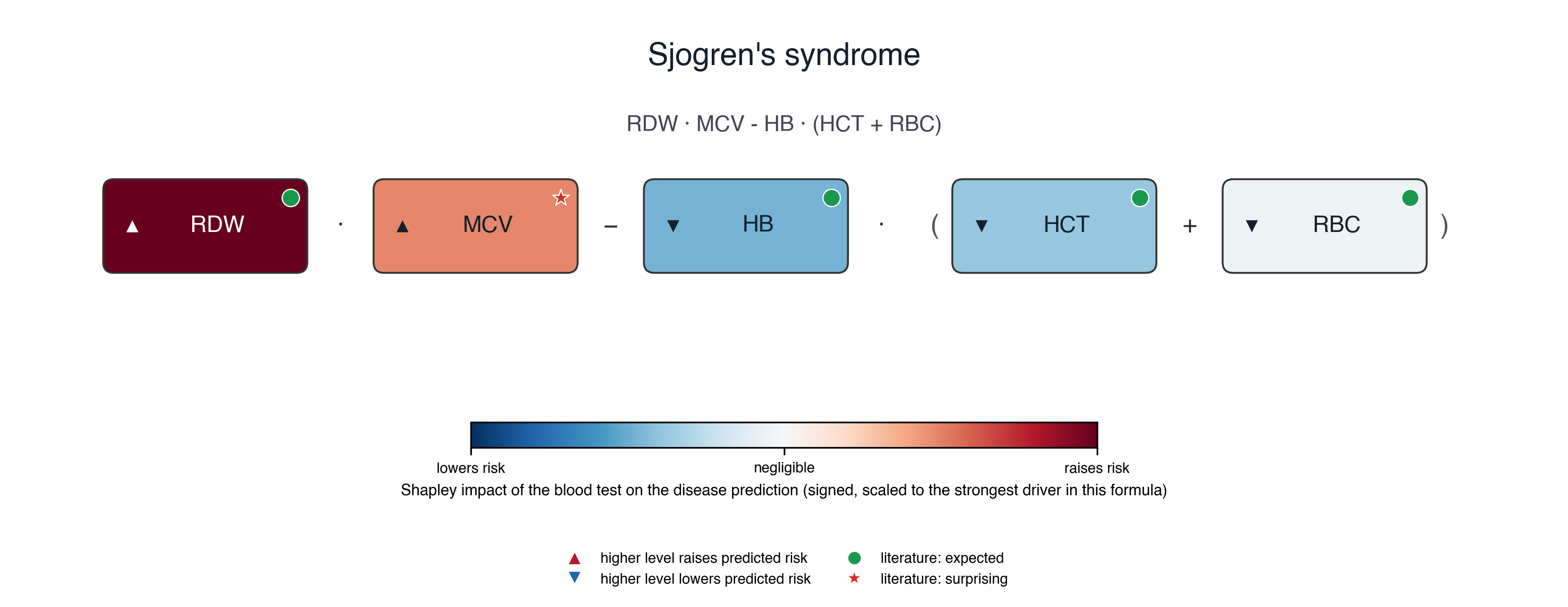}
\end{center}
{\footnotesize Each blood-test leaf is coloured by its Shapley impact on the disease prediction (red = raises risk, blue = lowers it; triangle = direction; circle = expected, star = surprising).\par}
\vspace{2pt}
\textbf{Biomarker:} RDW \ensuremath{\cdot} MCV - HB \ensuremath{\cdot} (HCT + RBC) $\cdot$ out-of-sample AUC (MIMIC) $\approx$ 0.5964\par
\textbf{RDW} (numerator). Higher RDW raises the predicted risk --- \textbf{expected}, consistent with RDW being elevated in Sjogren's syndrome (high confidence). Primary Sjogren's syndrome is a chronic systemic autoimmune/inflammatory disease. \cite{hu2014red}\par
\textbf{MCV} (numerator). Higher MCV raises the predicted risk --- \textbf{surprising}, since MCV is typically not consistently linked to Sjogren's syndrome (high confidence). In primary Sjogren's syndrome (pSS) the dominant red-cell abnormality is anemia of chronic disease (ACD), driven by cytokine-mediated (IL-6/hepcidin) iron sequestration and blunted erythropoiesis. \cite{yldz2021haematologic}\par
\textbf{HB} (numerator). Higher HB lowers the predicted risk --- \textbf{expected}, consistent with HB being reduced in Sjogren's syndrome (high confidence). In primary Sjogren's syndrome (pSS), hemoglobin is characteristically reduced. \cite{zhou2010clinical}\par
\textbf{HCT} (numerator). Higher HCT lowers the predicted risk --- \textbf{expected}, consistent with HCT being reduced in Sjogren's syndrome (high confidence). Reduced HCT in Sjogren's reflects anemia driven primarily by anemia of chronic disease (inflammatory cytokine/hepcidin-mediated iron sequestration and suppressed erythropoiesis), plus autoimmune hemolytic anemia (anti-erythrocyte antibodies, positive Coombs), immune-mediated bone-marrow involvement, and nutritional/renal contributions (e.g., associated autoimmune gastritis with B12 deficiency, renal tubular disease). \cite{zhou2010clinical}\par
\textbf{RBC} (numerator). Higher RBC lowers the predicted risk --- \textbf{expected}, consistent with RBC being reduced in Sjogren's syndrome (high confidence). In primary Sjogren's syndrome the dominant red-cell abnormality is anemia of chronic disease (ACD): chronic immune-mediated inflammation drives IL-6 and hepcidin upregulation, which sequesters iron in macrophages, blunts erythropoietin response, and shortens red-cell survival, lowering hemoglobin, hematocrit, and the RBC count. \cite{baimpa2009hematologic}\par
\vspace{2pt}
\clearpage
\section*{Systemic sclerosis}
\addcontentsline{toc}{section}{Systemic sclerosis (7101)}
\begin{center}
\includegraphics[width=0.88\textwidth]{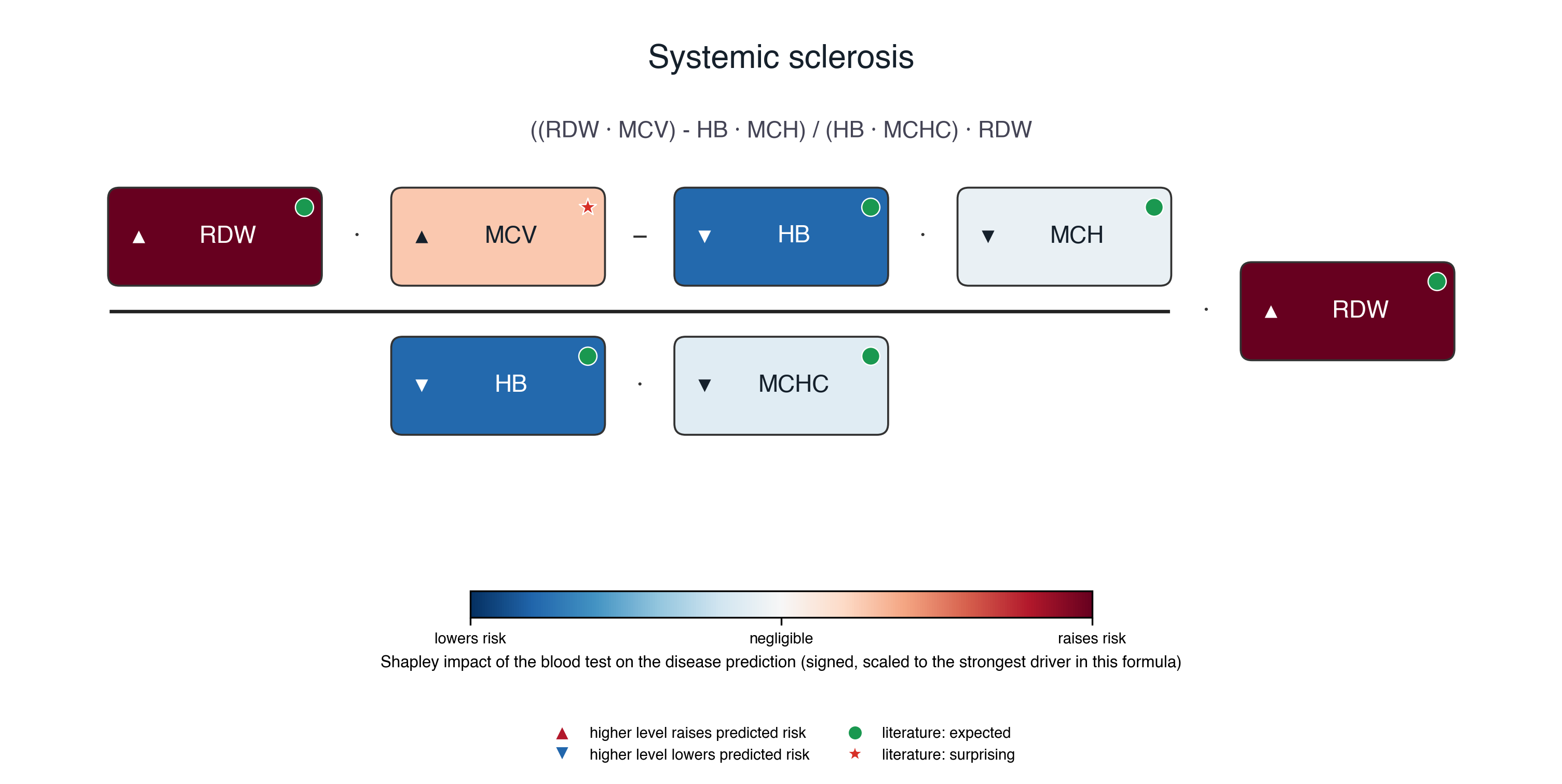}
\end{center}
{\footnotesize Each blood-test leaf is coloured by its Shapley impact on the disease prediction (red = raises risk, blue = lowers it; triangle = direction; circle = expected, star = surprising).\par}
\vspace{2pt}
\textbf{Biomarker:} ((RDW \ensuremath{\cdot} MCV) - HB \ensuremath{\cdot} MCH) / (HB \ensuremath{\cdot} MCHC) \ensuremath{\cdot} RDW $\cdot$ out-of-sample AUC (MIMIC) $\approx$ 0.5875\par
\textbf{RDW} (numerator). Higher RDW raises the predicted risk --- \textbf{expected}, consistent with RDW being elevated in Systemic sclerosis (high confidence). RDW (anisocytosis) rises in systemic sclerosis as a downstream readout of the disease's core pathophysiology: chronic systemic inflammation (RDW correlates positively with ESR and CRP, and inflammatory cytokines such as IL-6/TNF suppress erythropoietin response and impair iron metabolism, releasing heterogeneously sized reticulocytes), oxidative stress, and ineffective erythropoiesis. \cite{farkas2014clinical}\par
\textbf{HB} (numerator+denominator). Higher HB lowers the predicted risk --- \textbf{expected}, consistent with HB being reduced in Systemic sclerosis (high confidence). In systemic sclerosis (SSc), hemoglobin is characteristically reduced rather than elevated. \cite{gachet2024prevalence}\par
\textbf{MCV} (numerator). Higher MCV raises the predicted risk --- \textbf{surprising}, since MCV is typically reduced in Systemic sclerosis (high confidence). In systemic sclerosis the dominant, best-characterized red-cell-size abnormality is MICROCYTIC (low-MCV) anemia from chronic gastrointestinal blood loss leading to iron deficiency. \cite{gachet2024prevalence}\par
\textbf{MCHC} (denominator). Higher MCHC lowers the predicted risk --- \textbf{expected}, consistent with MCHC being reduced in Systemic sclerosis (medium confidence). Anemia is common in systemic sclerosis (SSc) and marks more severe disease. \cite{gachet2024prevalence}\par
\textbf{MCH} (numerator). Higher MCH lowers the predicted risk --- \textbf{expected}, consistent with MCH being reduced in Systemic sclerosis (medium confidence). MCH is the mass of hemoglobin per red cell; it falls in hypochromic anemia, classically iron-deficiency anemia (IDA). \cite{marie2008watermelon}\par
\vspace{2pt}

\clearpage
\section*{Ulcerative colitis}
\addcontentsline{toc}{section}{Ulcerative colitis (556)}
\begin{center}
\includegraphics[width=1.00\textwidth]{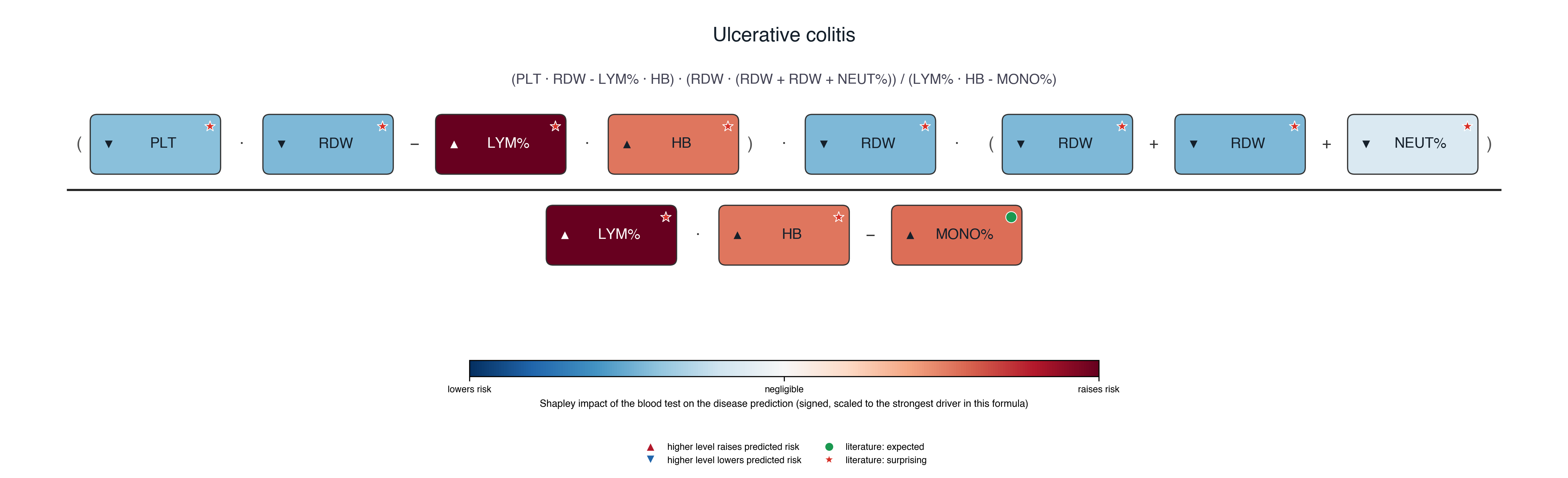}
\end{center}
{\footnotesize Each blood-test leaf is coloured by its Shapley impact on the disease prediction (red = raises risk, blue = lowers it; triangle = direction; circle = expected, star = surprising).\par}
\vspace{2pt}
\textbf{Biomarker:} (PLT \ensuremath{\cdot} RDW - LYM\% \ensuremath{\cdot} HB) \ensuremath{\cdot} (RDW \ensuremath{\cdot} (RDW + RDW + NEUT\%)) / (LYM\% \ensuremath{\cdot} HB - MONO\%) $\cdot$ out-of-sample AUC (MIMIC) $\approx$ 0.5715\par
\textbf{LYM\%} (numerator+denominator). Higher LYM\% raises the predicted risk --- \textbf{surprising}, since LYM\% is typically reduced in Ulcerative colitis (high confidence). In ulcerative colitis, acute mucosal inflammation drives a neutrophil-predominant systemic response (IL-6/STAT3, NLRP3/IL-1beta, Th17 expansion). \cite{ma2021application}\par
\textbf{MONO\%} (denominator). Higher MONO\% raises the predicted risk --- \textbf{expected}, consistent with MONO\% being elevated in Ulcerative colitis (high confidence). Ulcerative colitis is driven by innate immune activation. \cite{cherfane2015monocytosis}\par
\textbf{HB} (numerator+denominator). Higher HB raises the predicted risk --- \textbf{surprising}, since HB is typically reduced in Ulcerative colitis (high confidence). In ulcerative colitis, hemoglobin is characteristically LOWERED. \cite{filmann2014prevalence}\par
\textbf{RDW} (numerator). Higher RDW lowers the predicted risk --- \textbf{surprising}, since RDW is typically elevated in Ulcerative colitis (medium confidence). In ulcerative colitis, RDW (anisocytosis) is consistently raised by two overlapping mechanisms. (1) Anemia, predominantly iron-deficiency anemia from chronic micro/macroscopic blood loss across the ulcerated colonic mucosa plus malabsorption of iron/folate/B12, directly widens the RBC size distribution. (2) Chronic systemic inflammation: pro-inflammatory cytokines (IL-1, IL-6, TNF-alpha, IFN-gamma) impair erythropoietin production and marrow responsiveness, dysregulate iron metabolism (anemia of chronic disease), shorten RBC lifespan and produce heterogeneous, ineffective erythropoiesis. \cite{katsaros2020red}\par
\textbf{PLT} (numerator). Higher PLT lowers the predicted risk --- \textbf{surprising}, since PLT is typically elevated in Ulcerative colitis (high confidence). Platelet count is characteristically ELEVATED in ulcerative colitis (reactive/secondary thrombocytosis). \cite{talstad1973thrombocytosis}\par
\textbf{NEUT\%} (numerator). Higher NEUT\% lowers the predicted risk --- \textbf{surprising}, since NEUT\% is typically elevated in Ulcerative colitis (high confidence). Neutrophilic infiltration of the colonic mucosa is the defining histologic feature of ACTIVE ulcerative colitis: neutrophils migrate into the crypt epithelium (cryptitis) and lumen (crypt abscesses), and epithelial-derived chemokines (notably IL-8/CXCL8) drive this recruitment. \cite{omer2025diagnostic}\par
\vspace{2pt}
\clearpage
\section*{Crohn's disease}
\addcontentsline{toc}{section}{Crohn's disease (555)}
\begin{center}
\includegraphics[width=0.75\textwidth]{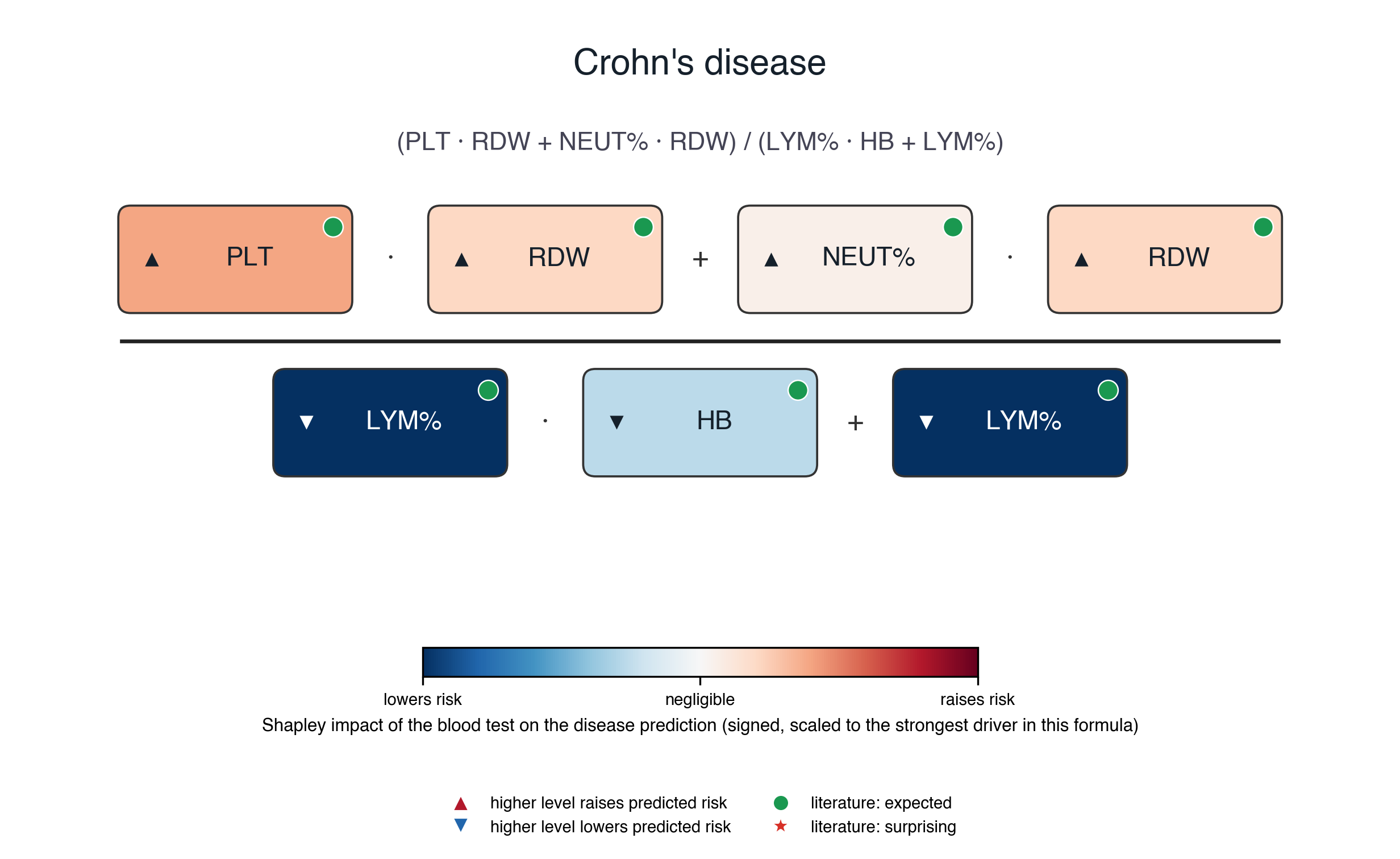}
\end{center}
{\footnotesize Each blood-test leaf is coloured by its Shapley impact on the disease prediction (red = raises risk, blue = lowers it; triangle = direction; circle = expected, star = surprising).\par}
\vspace{2pt}
\textbf{Biomarker:} (PLT \ensuremath{\cdot} RDW + NEUT\% \ensuremath{\cdot} RDW) / (LYM\% \ensuremath{\cdot} HB + LYM\%) $\cdot$ out-of-sample AUC (MIMIC) $\approx$ 0.5697\par
\textbf{LYM\%} (denominator). Higher LYM\% lowers the predicted risk --- \textbf{expected}, consistent with LYM\% being reduced in Crohn's disease (high confidence). Active Crohn's disease drives systemic inflammation: chronic immune activation expands the circulating neutrophil pool (demargination, IL-6/G-CSF-driven granulopoiesis) while the relative lymphocyte fraction falls (relative/absolute lymphopenia from cortisol-mediated redistribution, lymphocyte trafficking into inflamed gut mucosa, and chronic-inflammation-associated apoptosis). \cite{gao2015neutrophil}\par
\textbf{PLT} (numerator). Higher PLT raises the predicted risk --- \textbf{expected}, consistent with PLT being elevated in Crohn's disease (high confidence). Reactive (secondary) thrombocytosis is a well-recognized feature of active Crohn's disease and IBD generally. \cite{matowickakarna2016markers}\par
\textbf{HB} (denominator). Higher HB lowers the predicted risk --- \textbf{expected}, consistent with HB being reduced in Crohn's disease (high confidence). Anemia (low hemoglobin) is the most common systemic/extraintestinal complication of Crohn's disease, so hemoglobin is consistently REDUCED relative to controls. \cite{filmann2014prevalence}\par
\textbf{RDW} (numerator). Higher RDW raises the predicted risk --- \textbf{expected}, consistent with RDW being elevated in Crohn's disease (high confidence). RDW quantifies anisocytosis (heterogeneity of red-cell volume). \cite{hu2015value}\par
\textbf{NEUT\%} (numerator). Higher NEUT\% raises the predicted risk --- \textbf{expected}, consistent with NEUT\% being elevated in Crohn's disease (high confidence). Crohn's disease is a chronic transmural neutrophilic inflammatory disorder. \cite{he2025lymphocyte}\par
\vspace{2pt}

\clearpage
\section*{Multiple sclerosis}
\addcontentsline{toc}{section}{Multiple sclerosis (340)}
\begin{center}
\includegraphics[width=0.88\textwidth]{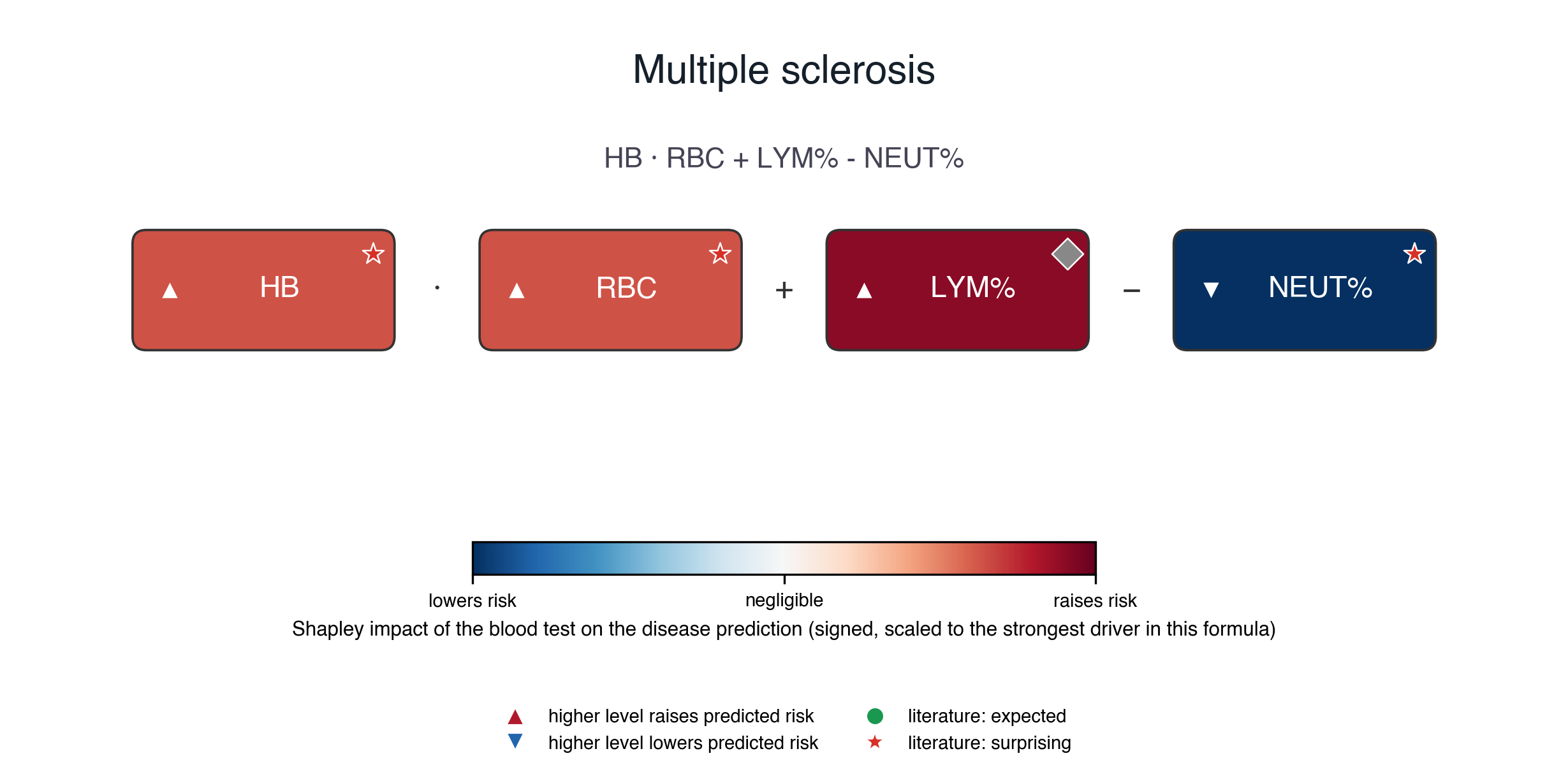}
\end{center}
{\footnotesize Each blood-test leaf is coloured by its Shapley impact on the disease prediction (red = raises risk, blue = lowers it; triangle = direction; circle = expected, star = surprising).\par}
\vspace{2pt}
\textbf{Biomarker:} HB \ensuremath{\cdot} RBC + LYM\% - NEUT\% $\cdot$ out-of-sample AUC (MIMIC) $\approx$ 0.5697\par
\textbf{NEUT\%} (numerator). Higher NEUT\% lowers the predicted risk --- \textbf{surprising}, since NEUT\% is typically elevated in Multiple sclerosis (medium confidence). In multiple sclerosis, low-grade systemic innate inflammation is part of the pathophysiology: cytokines/chemokines mobilize and prime neutrophils, which contribute to blood-brain-barrier disruption, neutrophil-extracellular-trap (NET) formation, and amplification of autoimmune CNS demyelination (corroborated in the EAE model). \cite{olsson2021neutrophil}\par
\textbf{LYM\%} (numerator). Higher LYM\% raises the predicted risk --- \textbf{unclear}, as the literature on LYM\% in Multiple sclerosis is limited (medium confidence). Two opposing strands exist. (1) Cross-sectional/prevalent-disease phenotype: MS involves myeloid/innate pro-inflammatory priming together with peripheral redistribution and sequestration of pathogenic lymphocytes into the CNS, so the circulating lymphocyte fraction is RELATIVELY reduced. \cite{olsson2021neutrophil}\par
\textbf{HB} (numerator). Higher HB raises the predicted risk --- \textbf{surprising}, since HB is typically reduced in Multiple sclerosis (medium confidence). In MS, the dominant published direction is reduced HB (anemia), driven by chronic immune-mediated inflammation (anemia of chronic disease with hepcidin-mediated iron sequestration), frequent B12/folate insufficiency (elevated homocysteine, megaloblastic features), reduced mobility/nutrition, and disease-modifying-therapy effects (e.g. natalizumab inhibiting the alpha4 integrin on erythroid precursors). \cite{koudriavtseva2015association}\par
\textbf{RBC} (numerator). Higher RBC raises the predicted risk --- \textbf{surprising}, since RBC is typically reduced in Multiple sclerosis (medium confidence). In MS, erythrocyte abnormalities are a recognized peripheral feature rather than an incidental finding. \cite{koudriavtseva2015association}\par
\vspace{2pt}
\clearpage
\section*{Type 1 diabetes}
\addcontentsline{toc}{section}{Type 1 diabetes (250)}
\begin{center}
\includegraphics[width=0.62\textwidth]{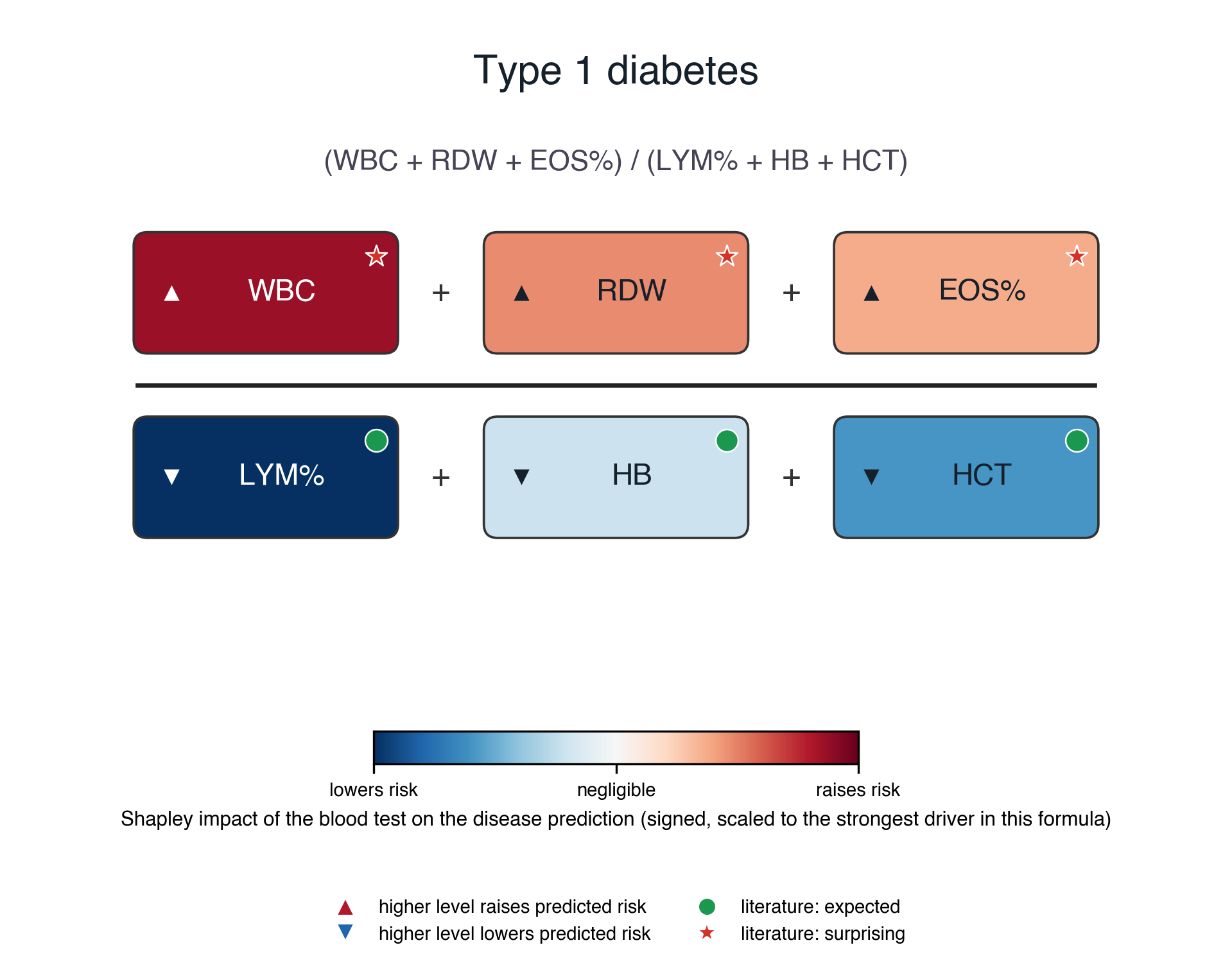}
\end{center}
{\footnotesize Each blood-test leaf is coloured by its Shapley impact on the disease prediction (red = raises risk, blue = lowers it; triangle = direction; circle = expected, star = surprising).\par}
\vspace{2pt}
\textbf{Biomarker:} (WBC + RDW + EOS\%) / (LYM\% + HB + HCT) $\cdot$ out-of-sample AUC (MIMIC) $\approx$ 0.5679\par
\textbf{LYM\%} (denominator). Higher LYM\% lowers the predicted risk --- \textbf{expected}, consistent with LYM\% being reduced in Type 1 diabetes (high confidence). T1D is a T-/B-cell-mediated autoimmune disease. \cite{salah2022leukocytic}\par
\textbf{WBC} (numerator). Higher WBC raises the predicted risk --- \textbf{surprising}, since WBC is typically reduced in Type 1 diabetes (high confidence). T1D is a T-cell-mediated autoimmune destruction of pancreatic beta cells, not a systemic myeloid-inflammatory disease, so a chronically elevated total WBC is not a recognized hallmark. \cite{salami2018reduction}\par
\textbf{HCT} (denominator). Higher HCT lowers the predicted risk --- \textbf{expected}, consistent with HCT being reduced in Type 1 diabetes (high confidence). In Type 1 diabetes, hematocrit (the packed red-cell fraction) trends LOWER than in non-diabetic individuals, primarily as a consequence of disease rather than a cause. \cite{thomas2004anemia}\par
\textbf{RDW} (numerator). Higher RDW raises the predicted risk --- \textbf{surprising}, since RDW is typically elevated in Type 1 diabetes (high confidence). In diabetes generally, RDW (erythrocyte volume heterogeneity) rises with chronic low-grade inflammation, oxidative stress, and impaired/ineffective erythropoiesis. \cite{zhang2018red}\par
\textbf{EOS\%} (numerator). Higher EOS\% raises the predicted risk --- \textbf{surprising}, since EOS\% is typically reduced in Type 1 diabetes (high confidence). In the observable CBC phenotype, eosinophils tend to be REDUCED in type 1 diabetes and in acute hyperglycemic stress. \cite{bambo2024changes}\par
\textbf{HB} (denominator). Higher HB lowers the predicted risk --- \textbf{expected}, consistent with HB being reduced in Type 1 diabetes (low confidence). Hemoglobin tends to be modestly LOWER in type 1 diabetes than in healthy peers via several converging mechanisms: (1) reduced renal erythropoietin production from incipient/established diabetic nephropathy --- the dominant driver in adults, where anemia tracks albuminuria and renal impairment (Thomas 2004); (2) iron-restricted erythropoiesis with elevated hepcidin and reduced TIBC, plus frank iron-deficiency anemia that is most pronounced around the time of diagnosis (Rusak 2018; Wojciak 2014); (3) chronic low-grade inflammation; and (4) T1D-enriched autoimmune comorbidities that cause anemia --- autoimmune/pernicious gastritis (B12 deficiency, \textasciitilde{}2.6-10\% prevalence) and celiac disease (iron/folate malabsorption), both substantially more common in T1D (Kahaly 2016). \cite{tihikapidi2021assessment}\par
\vspace{2pt}
\clearpage
\section*{Celiac disease}
\addcontentsline{toc}{section}{Celiac disease (5790)}
\begin{center}
\includegraphics[width=0.62\textwidth]{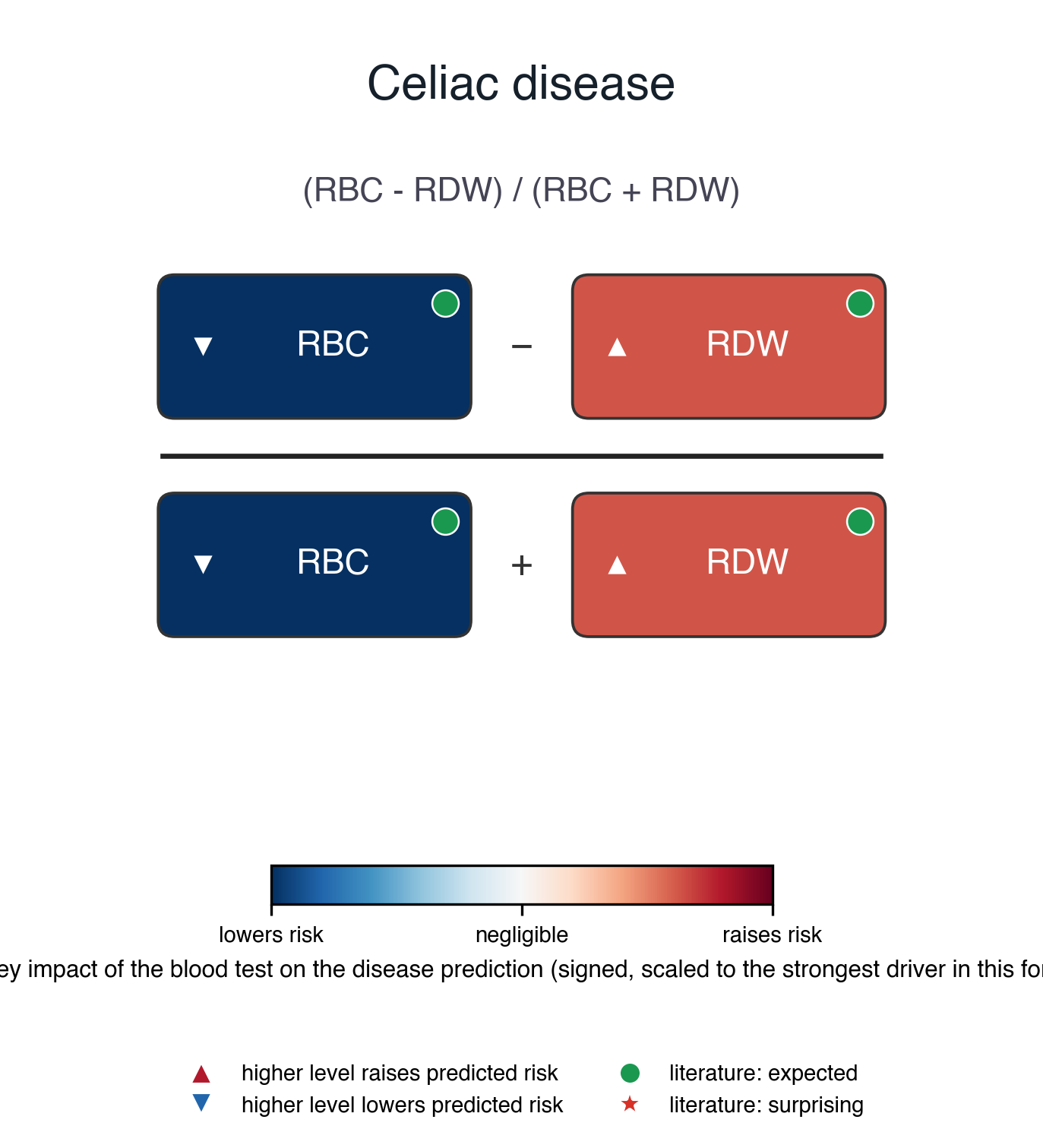}
\end{center}
{\footnotesize Each blood-test leaf is coloured by its Shapley impact on the disease prediction (red = raises risk, blue = lowers it; triangle = direction; circle = expected, star = surprising).\par}
\vspace{2pt}
\textbf{Biomarker:} (RBC - RDW) / (RBC + RDW) $\cdot$ out-of-sample AUC (MIMIC) $\approx$ 0.5569\par
\textbf{RBC} (numerator+denominator). Higher RBC lowers the predicted risk --- \textbf{expected}, consistent with RBC being reduced in Celiac disease (high confidence). In active celiac disease, immune-mediated duodenal/proximal-small-bowel villous atrophy impairs absorption of iron (and to a lesser degree folate and vitamin B12), producing iron-deficiency (and sometimes mixed-nutritional or chronic-disease) anemia. \cite{mahadev2018prevalence}\par
\textbf{RDW} (numerator+denominator). Higher RDW raises the predicted risk --- \textbf{expected}, consistent with RDW being elevated in Celiac disease (high confidence). Celiac disease causes immune-mediated villous atrophy of the proximal small intestine, the principal site of iron (and folate) absorption. \cite{sategnaguidetti2002red}\par
\vspace{2pt}

\clearpage
\bibliographystyle{naturemag}
\bibliography{biomarker_explainer_refs}